    \documentclass[pdflatex,sn-mathphys-num, iicol]{sn-jnl}

\usepackage{graphicx}%
\usepackage{multirow}%
\usepackage{amsmath,amssymb,amsfonts}%
\usepackage{amsthm}%
\usepackage{mathrsfs}%
\usepackage[title]{appendix}%
\usepackage{xcolor}%
\usepackage{textcomp}%
\usepackage{manyfoot}%
\usepackage{booktabs}%
\usepackage{algorithm}%
\usepackage{algorithmicx}%
\usepackage{algpseudocode}%
\usepackage{listings}%
\usepackage{makecell}
\usepackage{stfloats}
\usepackage{balance}
\usepackage{tabularx} 
\usepackage{subcaption}
\usepackage{cuted}
\usepackage{caption} 
\usepackage{csquotes}
\usepackage{hyperref}
\usepackage{tikz}
\usepackage{etoolbox}

\theoremstyle{thmstyleone}%
\theoremstyle{thmstyletwo}%

\theoremstyle{thmstylethree}%

\begin{document}

\title[Article Title]{\Large\boldmath Shapes are not enough: 
            CONSERVAttack and its use for finding vulnerabilities and uncertainties in machine learning applications}

\author[1]{\fnm{Philip} \sur{Bechtle}}

\author[2]{\fnm{Lucie} \sur{Flek}}

\author[3]{\fnm{Philipp Alexander} \sur{Jung}}

\author[2]{\fnm{Akbar} \sur{Karimi}}

\author*[1]{\fnm{Timo} \sur{Saala}}\email{tsaala@uni-bonn.de}

\author[3]{\fnm{Alexander} \sur{Schmidt}}

\author[1]{\fnm{Matthias} \sur{Schott}}

\author[4]{\fnm{Philipp} \sur{Soldin}}

\author[4]{\fnm{Christopher} \sur{Wiebusch}}

\author[3]{\fnm{Ulrich} \sur{Willemsen}}

\affil[1]{\orgdiv{Institute of Physics}, \orgname{University of Bonn}, \orgaddress{\country{Germany}}}

\affil[2]{\orgdiv{Bonn-Aachen Institute of Technology}, \orgname{University of Bonn}, \orgaddress{\country{Germany}}}

\affil[3]{\orgdiv{Physics Institute III A}, \orgname{RWTH Aachen University}, \orgaddress{\country{Germany}}}

\affil[4]{\orgdiv{Physics Institute III B}, \orgname{RWTH Aachen University}, \orgaddress{\country{Germany}}}

\abstract{In High Energy Physics, as in many other fields of science, the application of machine learning techniques has been crucial in advancing our understanding of fundamental phenomena. Increasingly, deep learning models are applied to analyze both simulated and experimental data. In most experiments, a rigorous regime of testing for physically motivated systematic uncertainties is in place. The numerical evaluation of these tests for differences between the data on the one side and simulations on the other side quantifies the effect of potential sources of mismodelling on the machine learning output. In addition, thorough comparisons of marginal distributions and (linear) feature correlations between data and simulation in \enquote{control regions} are applied. However, the guidance by physical motivation, and the need to constrain comparisons to specific regions, does not guarantee that all possible sources of deviations have been accounted for. 
We therefore propose a new adversarial attack -- the CONSERVAttack -- designed to exploit the remaining space of hypothetical deviations between simulation and data after the above mentioned tests. The resulting adversarial perturbations are consistent within the uncertainty bounds---evading standard validation checks---while successfully fooling the underlying model. We further propose strategies to mitigate such vulnerabilities and argue that robustness to adversarial effects must be considered when interpreting results from deep learning in particle physics.}


\maketitle

\section{Introduction}

Deep learning has established itself as a powerful tool across a wide range of scientific domains, including medicine \cite{10.1093/bib/bbx044}, autonomous driving \cite{10258330}, and, more recently, the development of general-purpose Large Language Models \cite{zhao2025surveylargelanguagemodels}. High Energy Physics (HEP) is no exception: While machine learning approaches---such as simple multilayer perceptrons---have been used in the field for decades, modern deep neural networks are increasingly employed in generative tasks (e.g., detector simulation), regression tasks (e.g., event reconstruction), and classification tasks (e.g., event selection and real-time triggering at the Large Hadron Collider (LHC)).

Despite these successes, applying deep learning in precision-driven fields like HEP presents significant challenges. Traditionally, analyses rely on well-defined statistical frameworks to estimate uncertainties by quantifying statistical variations and performing systematic studies of potential mismodellings in the simulated data relative to real observations. The numerical treatment of these uncertainties is included in the statistical frameworks used for model parameter fitting and for the calculation of confidence intervals~\cite{Cowan:2010js}. These uncertainties are evaluated in hundreds of separate studies, each probing different \emph{physical} sources of discrepancy, and are cross-checked through comparisons between data and simulation in carefully designed \enquote{control} and \enquote{validation} regions of the data's feature space, as well as through independent analyses. The discovery of the Higgs boson represents a famous and successful example of the thorough application of these techniques~\cite{ATLAS:2012yve,CMS:2012qbp}. Marginal distributions of features and their linear pairwise correlations---for both real and simulated events---are routinely examined and validated to ensure accurate modelling and well-quantified uncertainties. However, these cross-checks rarely go beyond the inspection of marginals, correlations, and possibly a limited number of two-dimensional distributions. In addition, validation regions are typically used to compare the output distribution of a classifier or regressor between data and simulation.

Neural networks often rely on high-dimensional and non-linear correlations in both data and simulation, making a systematic validation of their predictions using only the techniques described above challenging. Even when substantial effort is invested in producing accurate simulated data for training, the corresponding validation procedures remain limited. In most cases, they rely on comparisons of marginal feature distributions and pairwise correlations between simulated and real events. These checks do not probe the full complexity of the models' decision boundaries.

In this work, we demonstrate that this commonly used validation strategy can potentially become insufficient once we move beyond explicitly studied, physically motivated deviations between data and simulation. As soon as additional hypothetical sources of mismodeling---whether they are forgotten, overlooked, or entirely unknown---are allowed, and once we acknowledge that the \enquote{signal region}, to which a classifier or regression model is ultimately applied, may be affected by different sources of discrepancy than the control or validation regions, the traditional validation procedure may no longer provide appropriate coverage. To quantify the potential impact of such effects, we construct adversarial perturbations that modify simulated events in a targeted manner, inducing substantial degradation in the performance of downstream deep learning models while keeping all marginal distributions and inter-feature correlations well withing their expected statistical uncertainties. Due to these perturbations remaining invisible to standard validation techniques, they represent a previously unexplored source of systematic uncertainty. The ability to generate such undetectable adversaries thus provides a new means of estimating an upper bound on a model's systematic vulnerability.

The techniques presented here are used to propose a workflow that estimates the maximum susceptibility of a network to these adversarial perturbations, while also outlining steps to reduce the resulting uncertainties to, or below, the level associated with known physically motivated sources. In this workflow, a final result indicating that the model's susceptibility is within the range of established physical uncertainties would suggest that no additional uncertainty from adversarial effects needs to be considered, since it can be assumed that the physically motivated systematics already cover the uncertainty from adversarials. Conversely, if the proposed workflow yields a final fooling ratio exceeding the magnitude of the physically motivated uncertainties, this would prompt either a review for potential omissions in the physical uncertainty estimates or the assignment of an additional uncertainty to account for misclassification due to adversaries of unknown origin.

We further investigate whether---in some cases---discrepancies between model performance on simulated and real events, as sometimes observed in analyses, may be partially explained by the presence of adversarially structured yet statistically consistent events in simulation.

Beyond assessing vulnerabilities and remaining uncertainties, we  demonstrate that adversarial events can be leveraged to improve model performance, even on unperturbed inputs. Specifically, we apply this attack as a form of data augmentation in low-data regimes.

Finally, we explore two adversarial defense strategies aimed at reducing the susceptibility of deep learning models to such attacks. The first, adversarial training, involves augmenting the training set with adversarial events. The second involves constructing an Adversarial Detector network, optimized to distinguish between clean and adversarially perturbed events. In both cases, we show that it is possible to improve the models robustness against these adversaries---consequently lowering the estimation of the uncertainty upper-bound. Additionally, we perform a  detailed analysis of the Adversarial Detector's behavior on both simulated and real HEP data, investigating whether certain simulated events exhibit adversarial behavior and how well the detector generalizes to unseen real events.


\section{Related Work}

Adversarial attacks and defenses are well-established fields of research across a variety of contexts \cite{10510296}, and have been particularly extensively studied in security-critical applications. Many sophisticated methods exist for constructing adversarial examples across diverse types of data, including images \cite{goodfellow2015explainingharnessingadversarialexamples, madry2019deeplearningmodelsresistant}, tabular data \cite{cartella2021adversarialattackstabulardata}, and even natural language \cite{zhang2019adversarialattacksdeeplearning}. However, these approaches do not necessarily address the specific requirements and constraints relevant to High Energy Physics (HEP).

Within HEP, research on adversarial attacks and, more broadly, adversarial deep learning remains relatively limited \cite{Flek2025Enforcing, flek2025minifoolphysicsconstraintawareminimizerbased}. In this work, we propose an adversarial attack specifically designed to address these domain-specific needs. We study the behavior of this attack in a general adversarial deep learning context, as well as its implications for downstream HEP applications and prior event simulations. Furthermore, we discuss the consequences of such attacks on uncertainty estimation in deep learning tasks and simulations within HEP, two areas that have been studied extensively. Nonetheless, we argue that the approach we use here to analyze model uncertainties has not yet been explored sufficiently.

Another line of work addressing robustness to distributional shifts is Distributionally Robust Optimization (DRO) \cite{Rahimian_2022}. In DRO, models are trained to minimize the worst-case loss over an uncertainty set of distributions surrounding the empirical training distribution. These uncertainty sets are typically defined through divergence measures, such as Wasserstein or $f$-divergence balls, allowing the optimization procedure to consider adversarial reweightings or shifts of the data distribution during training. Conceptually, this shares similarities with the adversarial perspective adopted in this work, as both aim to reason about worst-case deviations from the nominal data distribution. However, the assumptions and objectives differ substantially. In most DRO formulations, the distributions considered during optimization are not constrained to preserve specific statistical properties of the dataset, such as marginal feature distributions or pairwise linear correlations. As a result, the adversarial distributions explored by DRO may deviate in ways that would be easily detectable by the validation procedures commonly used in High Energy Physics analyses. In contrast, the perturbations constructed in this work are explicitly constrained to remain consistent with the statistical checks routinely performed in HEP, including agreement of marginal distributions and linear feature correlations within their expected uncertainties. Our approach therefore focuses on identifying adversarial deviations that would evade standard validation strategies, providing a complementary perspective to distributionally robust training methods.

\section{Method: CONSERVAttack}

The goal of this attack is to construct adversarial perturbations that remain physically invisible under analyses commonly applied in High Energy Physics (HEP). To achieve this, we must ensure that neither the marginal distributions of the input features nor the correlations between them are significantly altered. This requirement motivates a shift in perspective compared to other adversarial attack design. While most attacks typically seek to constrain the perturbation on a per-event basis (e.g., by minimizing the $L_\infty$ norm of each modified input), our method instead constrains changes at the dataset level, such as over an entire test set.

Similar to state-of-the-art adversarial attacks, such as Projected Gradient Descent (PGD), we aim to optimize a min-max problem. We want to maximize a goal function---such as reaching misclassification---while simultaneously minimizing the perturbations applied to data. For this attack, this minimization step involves two separate constraints evaluated over a batch of inputs: preserving marginal features distributions and preserving the inter-feature correlations. The attack proceeds via an iterative, gradient-based, brute-force–style search over candidate perturbations.

More specifically, for each input we compute the gradient of the model's loss function with respect to the input at the final layer. We then discard the gradient magnitudes, retaining only its sign. Most adversarial attack algorithms use the sign of the gradient to apply either a single large, or many small perturbations to the input, we use it to generate a set of candidate perturbations $S_{cp}$ for each feature. 

To this end, we introduce two hyperparameters: the minimal perturbation magnitude \(p_{\text{min}} > 0\), which sets the smallest allowed change to an input feature, and a step size \(\epsilon_{\text{step}}\), which determines the spacing between candidate perturbations. For each candidate in \(S_{pc}\), we evaluate its impact on both the marginal feature distributions and the feature--feature correlations.

Alternatively, the candidate generation procedure can be overridden by explicitly specifying the number of candidates to be considered via the parameter \(num_C\). When this option is used, the attack generates \(num_C\) uniformly distributed candidate perturbations between the current feature value and the corresponding minimum or maximum global value of that feature---depending on the gradient sign. 

To quantify changes in the marginal distributions, we compute the Jensen–Shannon Distance ($D_{\mathrm{JS}}$), defined as

\begin{align}
D_{\mathrm{JS}} := \sqrt{\dfrac{D(\Vec{p} \Vert \Vec{m}) + D(\Vec{q} \Vert \Vec{m})}{2}}, 
\label{eq:jsd}
\end{align}

where $D$ is the Kullback-Leibler divergence. Here, $\Vec{p}$ and $\Vec{q}$ are the normalized approximate probability distributions (estimated using histograms) and $\Vec{m}$ is their pointwise mean. 

To quantify changes in correlations, we define $\Delta_F$ as the relative Frobenius norm difference between the correlation matrices of the clean and perturbed inputs:

\begin{align}
\Delta_F :=
\frac{\lVert C_\text{adv} - C_\text{clean} \rVert_F}
{\lVert C_\text{clean} \rVert_F},
\label{eq:FN}
\end{align}

where $C_\text{clean}$ and $C_\text{adv}$ denote the correlation matrices of the original and perturbed inputs, respectively. The normalization by $\lVert C_\text{clean} \rVert_F$ expresses the change relative to the magnitude of the clean correlation matrix.

The Frobenius norm of a matrix $A$ is defined as

\begin{align}
\lVert A \rVert_F := \sqrt{\sum_{i=1}^{m} \sum_{j=1}^{n} a_{ij}^2},
\end{align}

where $A \in \mathbb{R}^{m \times n}$ and $a_{ij}$ denotes the entry in the $i$-th row and $j$-th column.

After computing $D_{\mathrm{JS}}$ and $\Delta_F$ for each candidate perturbation of each feature, we select the perturbation that minimizes a custom loss function.

\begin{align}
\mathcal{L} := \alpha D_{\mathrm{JS}} + \beta\Delta_F,
\label{eq:CustomLoss}
\end{align}

where $\alpha$ and $\beta$ hyperparameters controlling the relative weight of the two constraints. This is done for a pre-defined amount of iterations $n_{it}$. Moreover, setting \texttt{nc} = \texttt{True} allows the algorithm to skip an update entirely whenever no candidate modification satisfies the required bounds on the $D_{\mathrm{JS}}$ variation < $\max_{\mathrm{D_{\mathrm{JS}}}_t}$ and the Frobenius Norm change $\max_{\Delta \mathrm{FN}_t}$, both of which are user-configurable hyperparameters.

Moreover, throughout the optimization we maintain a dynamic boolean mask that tracks whether a given event already successfully fools the model under attack. By default, the flag $\mathrm{opt}_{af}$ (optimize already fooled events) is set to \texttt{False}. In this configuration, once an event achieves the adversarial objective (e.g., misclassification), it is excluded from further perturbations in subsequent iterations. However, these events are still included when computing dataset-level metrics such as the marginal feature distributions and inter-feature correlations, ensuring that the overall $D_{\mathrm{JS}}$ and $\Delta_F$ constraints continue to account for all data. This reduces unnecessary modifications to already successful adversarial examples while preserving accurate distributional monitoring.

Alternatively, setting $\mathrm{opt}_{af}$ to \texttt{True} keeps such events within the perturbation loop. In this case, the algorithm continues to evaluate and apply candidate perturbations to already successful adversarial examples, aiming to reduce their contribution to $D_{\mathrm{JS}}$ and $\Delta_F$ terms while maintaining their adversarial property. This optional refinement step can therefore produce adversarial examples that remain effective while exhibiting smaller deviations in the monitored distributional and correlation-based metrics.

Additional optimizations and implementation details---such as more computationally efficient procedures for evaluating $D_{\mathrm{JS}}$ and $\Delta_F$---are provided in the Appendix \ref{App:Optimization}

\section{Data Sets, Tasks and Classification Models Used for the Performance Evaluation}

We evaluate CONSERVAttack on two common particle physics tasks. The first task, referred to as Higgs, is based on the popular Higgs Boson Machine Learning Kaggle Challenge \cite{higgs-boson}. The second task is a jet-tagging problem, where the goal is to distinguish jets originating from a pair of top quarks (TTJets) from those originating from a pair of W bosons (WWJets). The datasets for this task were obtained from the CERN Open Data portal \cite{CERNOpenData}, more specifically using simulated samples from the 2012 CMS run \cite{TTJets, WWJets}. 

In both tasks, the objective is binary classification: TT vs. WW Jets, and signal (Higgs) vs. background processes. For the studies presented here, we apply simple Multi-Layer Perceptrons (MLP) to both tasks. For the TT vs. WW classification, the architecture of the deep learning model is based on the TopoDNN model \cite{10.21468/SciPostPhys.7.1.014}. 

The Higgs dataset contains a total of 30 continuous input features, and the corresponding MLP has a total of 42,163 trainable parameters. The TopoDNN-inspired model used for the jet-tagging task uses 87 continuous input features and has 59,263 trainable parameters. Hyperparameter choice, training details, as well as more technical details for both of these networks can be found in the Appendix \ref{App:DataNet}.

\section{Impact of CONSERVAttack}

Although adversarial robustness is generally not a security-critical concern in HEP, we still evaluate the attack against neural networks to assess its effectiveness. Our goal is to show that it is possible to construct adversarial inputs that, according to standard HEP simulation validation procedures, would be assigned to class A, while the downstream classifier misidentifies them as belonging to another class.

To evaluate the effectiveness of the attack, we consider both parts of the min–max optimization. 
For the maximization objective---that is, inducing a misclassification---we use the Fooling Ratio ($\mathcal{F}$), defined as

\begin{align}
\mathcal{F} := \frac{1}{N} \sum_{i=1}^{N} \mathbf{1}\!\left( f(x_i') \neq f(x_i) \right),
\end{align}

where $N$ denotes the total number of events, $f(\cdot)$ is the classifier, $x_i$ is the original input, and $x_i'$ is its adversarially perturbed counterpart. Here, $\mathbf{1}(\cdot)$ denotes the indicator function, which equals $1$ if the condition is true and $0$ otherwise. The Fooling Ratio therefore measures the fraction of inputs for which the predicted label changes under the adversarial perturbation. By construction, our attack updates each input along the gradient direction that increases the classifier loss, similar to standard PGD or FGSM approaches; as a result, the Fooling Ratio is implicitly optimized with each iteration without requiring an explicit term in the loss function.

To quantify the minimization component---here, the preservation of the dataset's statistical properties---we use the previously introduced Jensen-Shannon Distance (\ref{eq:jsd}) and $\Delta_F$ (\ref{eq:FN}). Both quantities are averaged over all input features, yielding a single value. Smaller values indicate a better match between the perturbed and original distributions or correlations.

\begin{figure*}[!htb]
     \centering
     \includegraphics[width=0.99\textwidth]{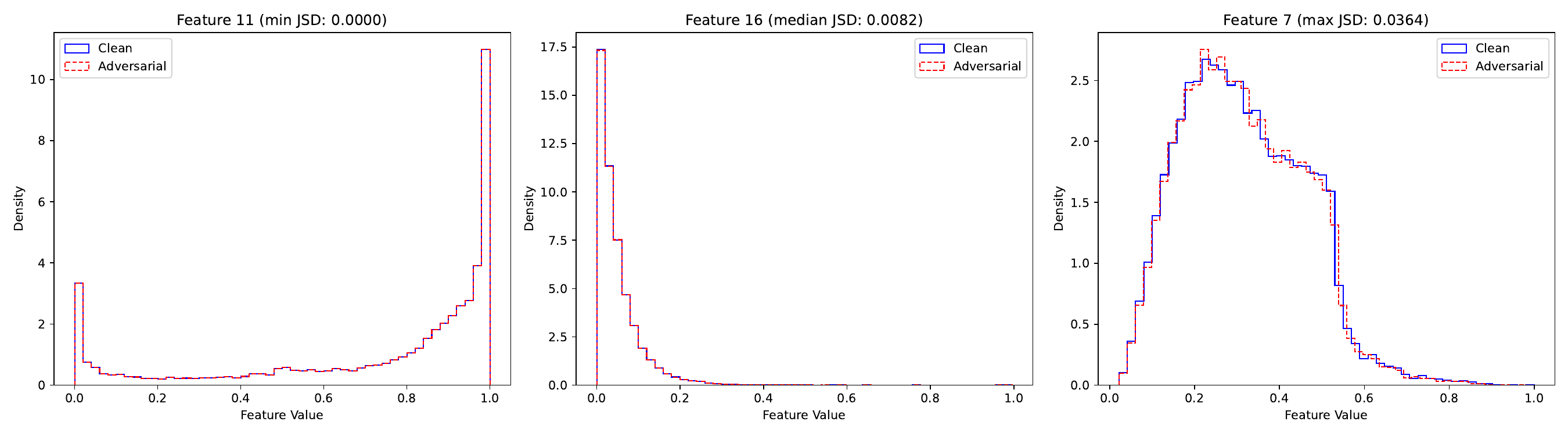}
     \caption{Comparison of three distinct distributions between clean and adversarial events in an attack setting. The plot on the left shows the feature resulting in the minimal change in Jensen Shannon distance, the one in the middle the feature corresponding to the median Jensen Shannon Distance, and the plot on the right the feature exhibiting the largest change in the Jensen Shannon Distance for the given attack run.}
     \label{fig:Higgs_JSD_Comp}
\end{figure*}

\begin{figure}[!htb]
     \centering
     \includegraphics[width=0.49\textwidth]{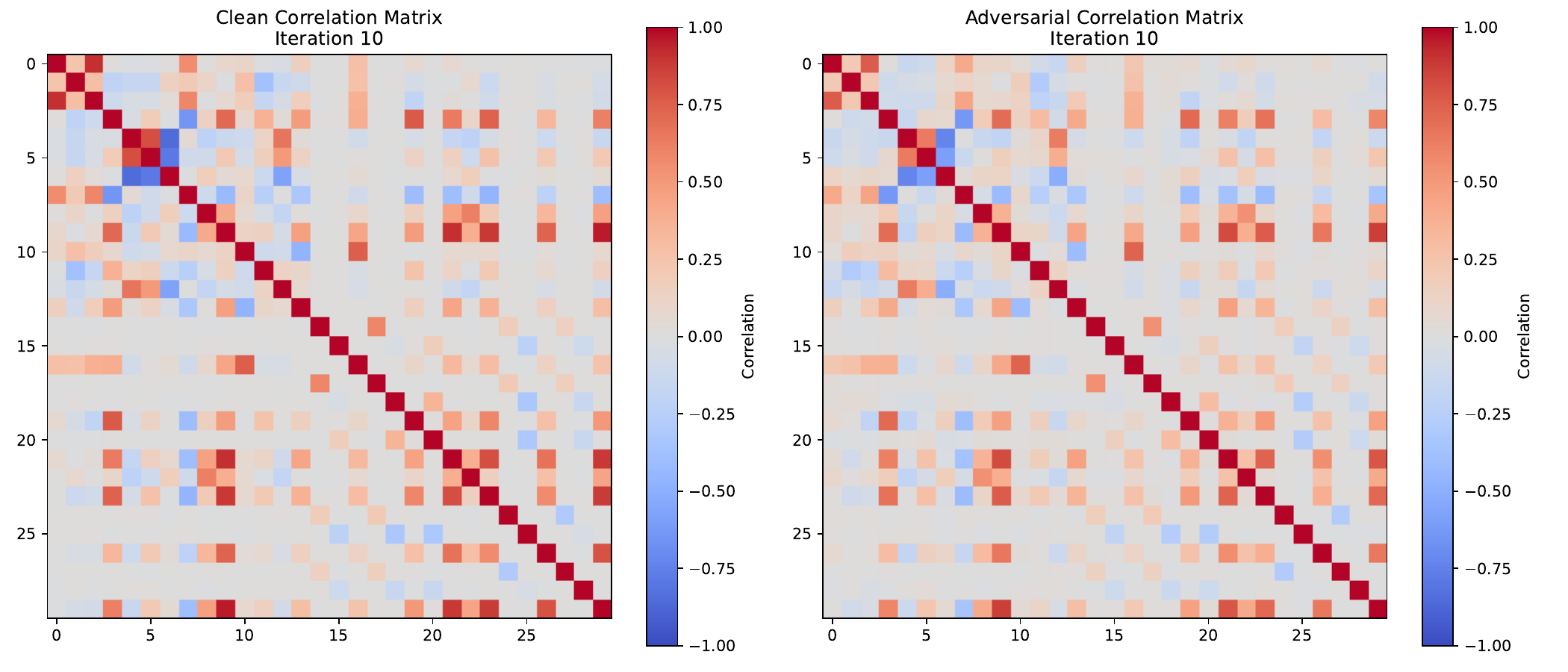}
     \caption{Comparison of the correlation matrices between the clean events (left) and adversarial events (right) resulting from the attack.}
     \label{fig:Higgs_FN_Comp}
\end{figure}

In contrast to the Fooling Ratio, these metrics are optimized explicitly via the combined loss function (\ref{eq:CustomLoss}) introduced in the Methods section, which penalizes deviations in marginal distributions and inter-feature correlations. While there is no set threshold for interpreting these metrics in terms of statistical significance, visual inspection of the resulting distributions and correlations allows for a good assessment. Example results for the Higgs task, where our attack achieves a Fooling Ratio of roughly 0.9, are shown in Figures \ref{fig:Higgs_JSD_Comp} and \ref{fig:Higgs_FN_Comp}.

The results indicate that, even in the worst case, the attack introduces only minor perturbations to the marginal distributions, and the resulting correlation matrices remain close to their original values.

To better assess the efficiency of the attack, we apply it ten times for each task. In each run, a new network is trained using identical hyperparameters but different random seeds for initialization. The attack parameters are fixed across all runs and are listed in Table \ref{tab:attack_params}.

\begin{table*}[htb!]
\caption{Attack parameters for both the Higgs and the TT vs. WW jet-tagging tasks. }
\centering
\begin{tabular}{|l|c|c|c|c|c|c|c|c|c|c|}
\hline
\textbf{Task} & \textbf{$p_{min}$} & \textbf{$\epsilon_{step}$} & \textbf{$n_{it}$} & \textbf{$\alpha$} & \textbf{$\beta$} & \textbf{$\max_{JSD_t}$} & \textbf{$\max_{\Delta_Ft}$} & \textbf{$nc$}\\
\hline
\hline
 Higgs & 0.0005 & 0.0005 & 10 & 4.0 & 1.0 & 0.006 & 0.0002 & True\\
\hline
 TTvsWW & 0.005 & 0.01 & 10 & 6.5 & 1.0 & 0.003 & 0.003 & True\\
\hline
\end{tabular}
\label{tab:attack_params}
\end{table*}

\begin{figure}[!htb]
     \centering
     \includegraphics[width=0.49\textwidth]{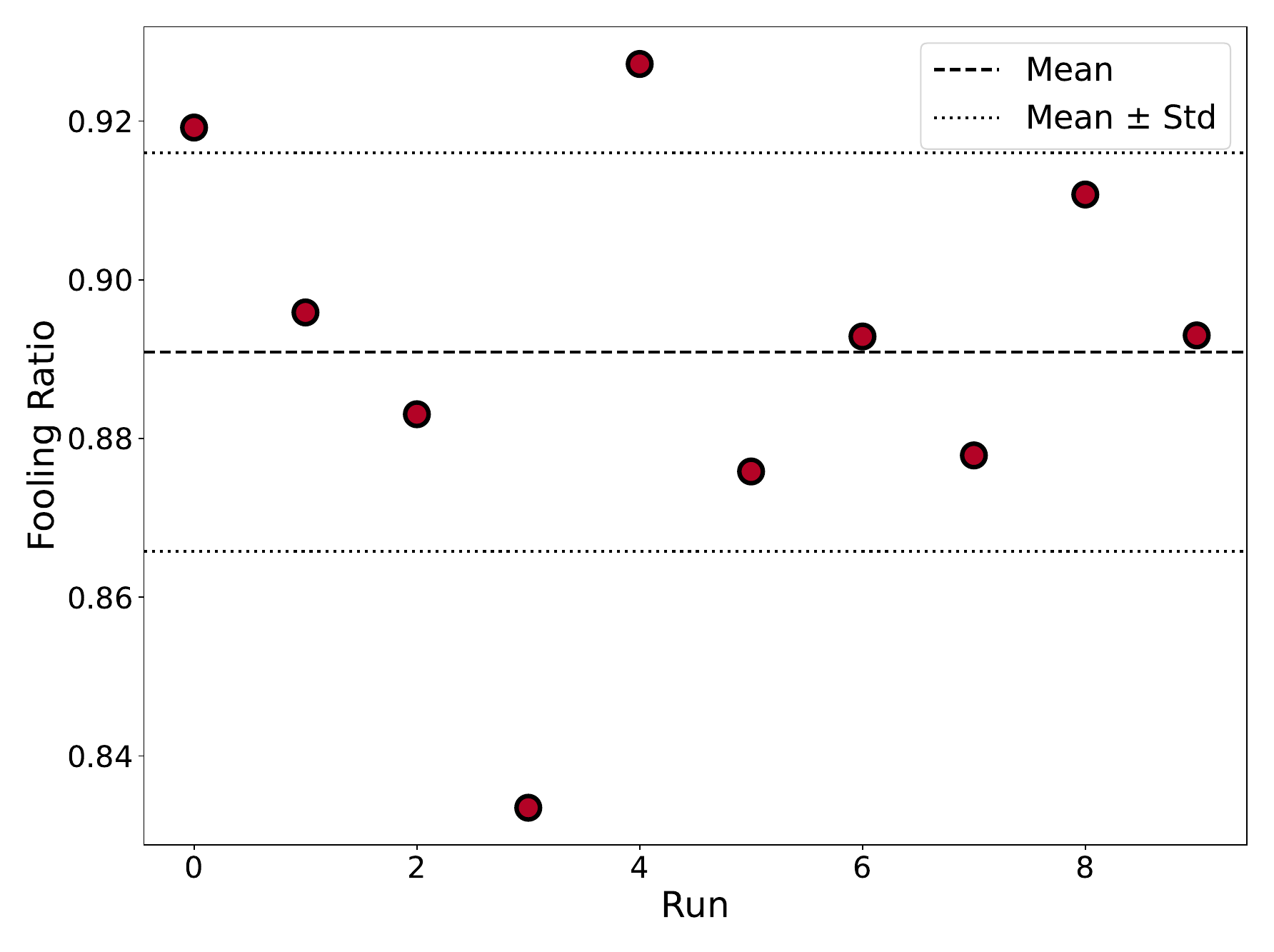}
     \caption{Fooling ratio of the adversarial attack on the Higgs dataset over 10 runs. The average performance of the attack is shown as a dashed line, and its standard deviation as a dotted line.}
     \label{fig:Higgs_FR_Unrestricted}
\end{figure}

First, we examine the maximization component of the attack---its ability to induce misclassification. The resulting Fooling Ratios for ten independent runs on the Higgs task are shown in Figure \ref{fig:Higgs_FR_Unrestricted}. 

\begin{figure}[!htb]
     \centering
     \includegraphics[width=0.49\textwidth]{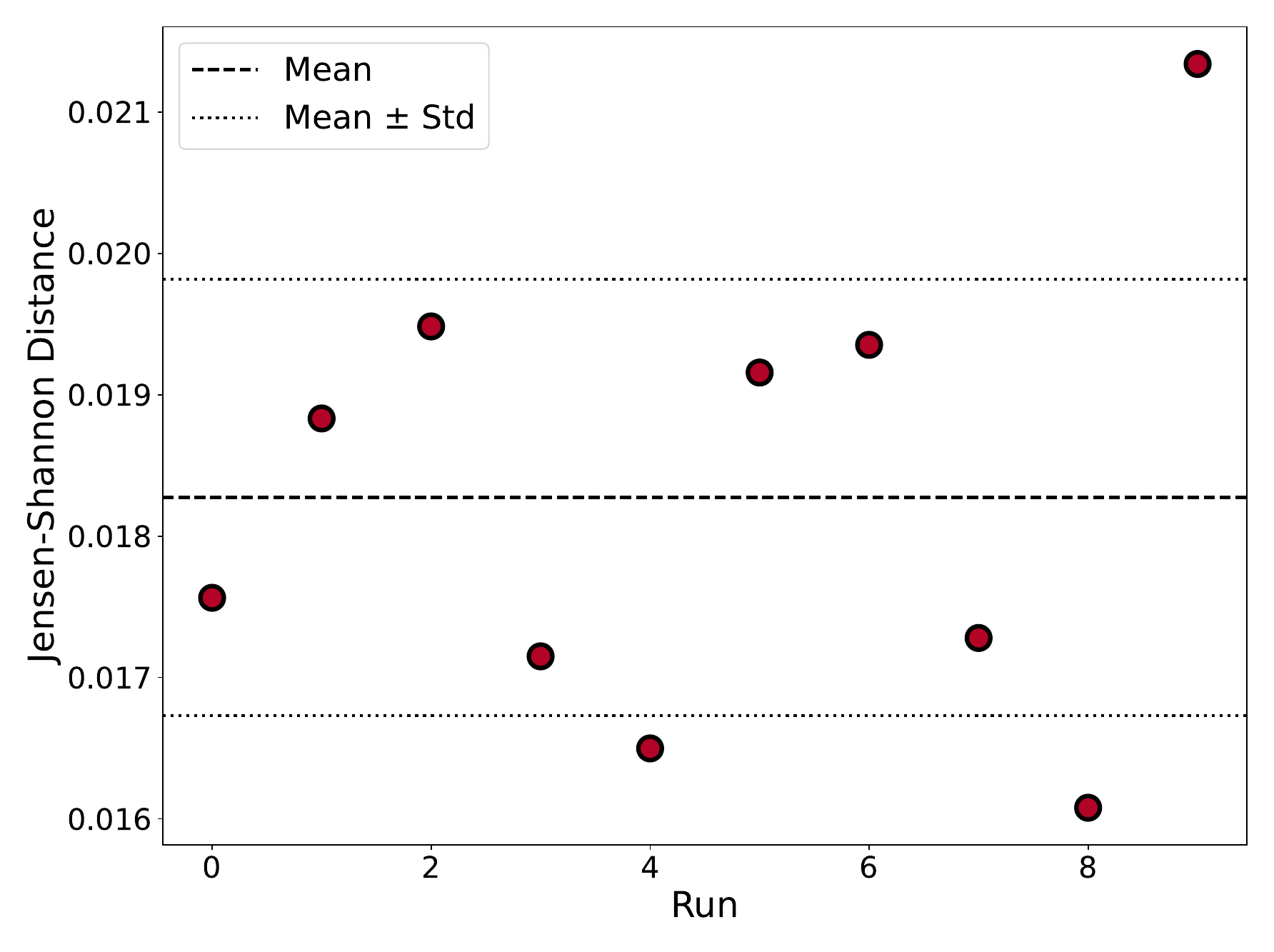}
     \caption{Average Jensen Shannon distance of the adversarial attack on the Higgs dataset over 10 runs. The average performance of the attack is shown as a dashed line, and its standard deviation as a dotted line.}
     \label{fig:Higgs_JSD_Unrestricted}
\end{figure}

\begin{figure}[!htb]
     \centering
     \includegraphics[width=0.49\textwidth]{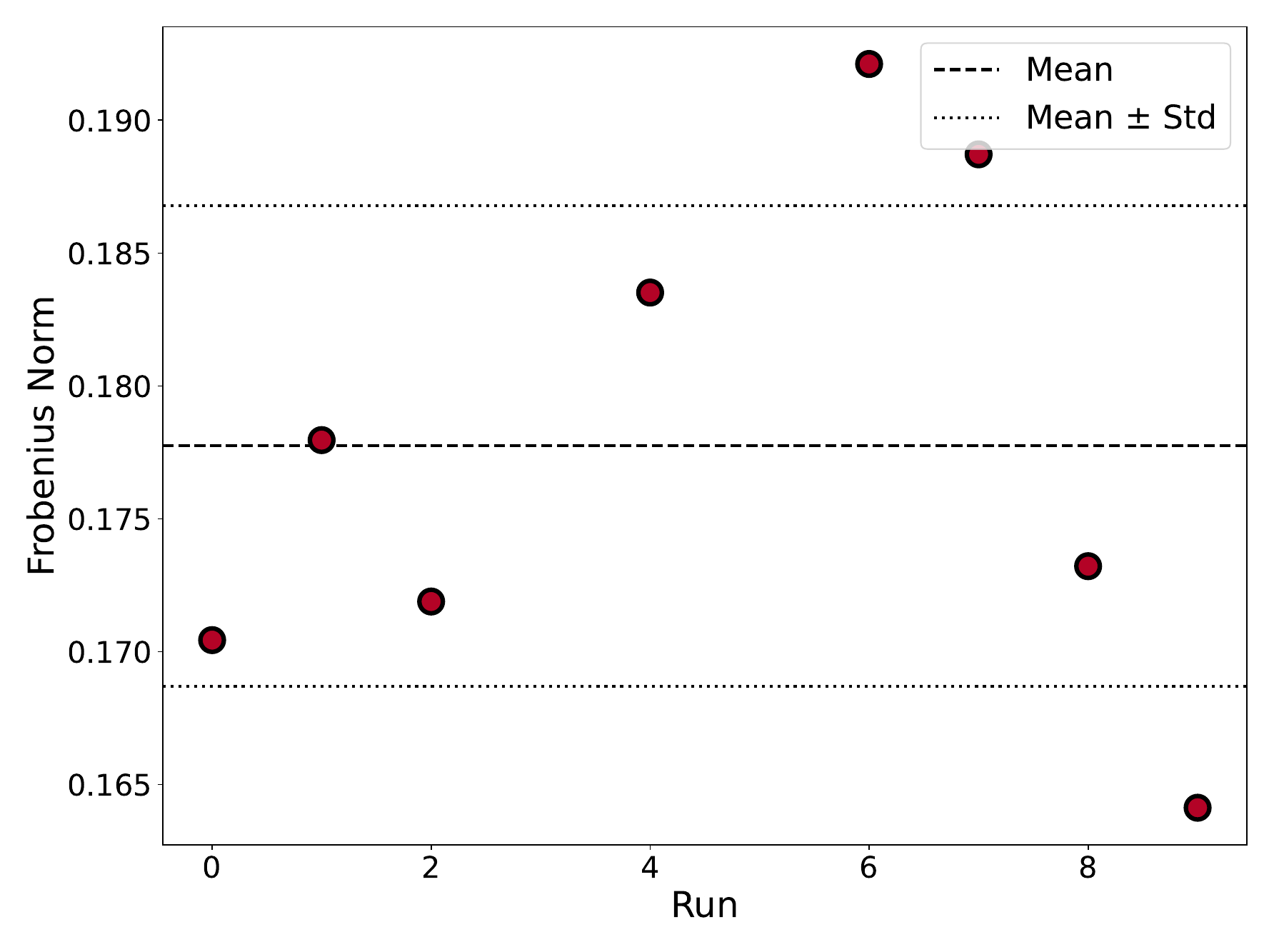}
     \caption{Average Frobenius Norm of the adversarial attack on the Higgs dataset over 10 runs. The average performance of the attack is shown as a dashed line, and its standard deviation as a dotted line.}
     \label{fig:Higgs_FN_Unrestricted}
\end{figure}

Across the ten runs, the average Fooling Ratio is 0.89, indicating that the attack consistently generates adversarial inputs that successfully alter the network's predictions. Next, we need to verify that these adversarial examples achieve this while minimally---ideally imperceptibly---modifying the underlying marginal distributions and feature correlations. The corresponding metrics, the Jensen-Shannon Distance and the Frobenius norm of the difference between correlation matrices, are shown in Figure \ref{fig:Higgs_JSD_Unrestricted} and \ref{fig:Higgs_FN_Unrestricted}.

For the Jensen-Shannon Distance, all but one run remain below 0.02, indicating a good match between the adversarial and clean distributions. Similarly, $\Delta_F$ remains below 0.2, demonstrating a low perturbation of the correlations. These results indicate that, across multiple independent runs, it is feasible to generate adversarial samples that induce misclassification while only minimally altering the marginal distributions and correlations.

The attack shows similar behavior on the TopoDNN network. Across ten independent runs, there, the average Fooling Ratio is 0.675, with an average $D_{\mathrm{JS}}$ of 0.045 and an average $\Delta_F$ of 0.182. Corresponding plots and additional results are provided in the Appendix \ref{App:AttackRes}.

Additionally, we performed a physically motivated variant of the attack on the Higgs dataset, in which the constrained attack is applied only to the control region, while fully unconstrained perturbations are applied to the complementary region. A more detailed description, along with the results and their discussion, is provided in Appendix \ref{App:CRAttack}.

\section{Data Augmentation using CONSERVAttack}
\label{sec:Augmentation}

\begin{figure*}[!htb]
     \centering
     \includegraphics[width=0.99\textwidth]{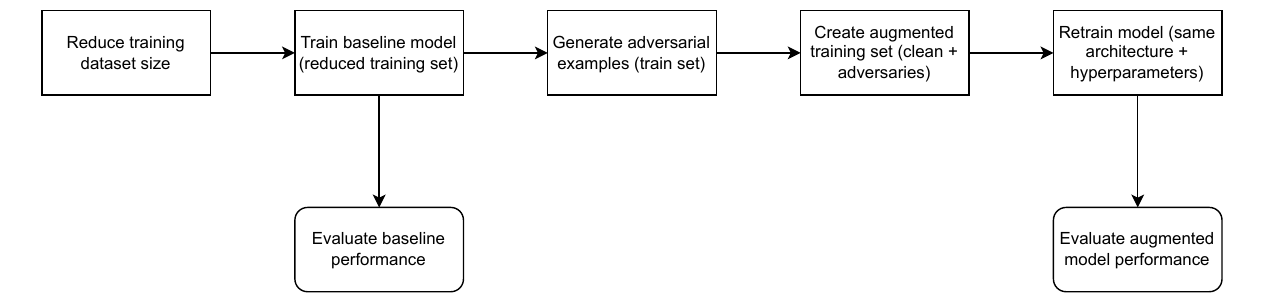}
     \caption{Flowchart showing the workflow used in the data augmentation pipeline.}
     \label{fig:AugmentationFlowchart}
\end{figure*}

Before addressing how to construct models that are robust to physically undetectable adversaries, we first explore the use of these adversarial attacks as a data-augmentation strategy. Prior work has shown that, in some settings, augmenting the training set with adversarial examples can improve the generalization capability of neural networks---not only against adversaries but also on clean inputs \cite{tsipras2018robustness}. However, in cases where a model is already saturated in its learning capacity (for example it has been trained on a large enough dataset), adversarial training can instead degrade the models performance on clean inputs.

In this study, we focus on the situation where the baseline model does not have enough training data to fully saturate. To force this scenario, we intentionally reduce the size of the training data for both the Higgs and the TT vs. WW tasks until the model's baseline performance noticeably deteriorates. For the Higgs network, the training set size is reduced from roughly 150.000 to 5.000 inputs, and for the TopoDNN-like model from about 280.000 to 70.000.

The augmentation procedure is as follows: First, the network is trained on the artificially reduced training set, resulting in an undertrained baseline model. This model is then used within our adversarial attack pipeline, but in contrast to the previous experiments, the adversaries are now generated from the training set. The resulting adversarial inputs are shuffled into the original training data; assuming a Fooling Ratio of 1.0, this yields at most one adversary per clean input, at most doubling the dataset size.

After generating the adversaries, the model's weights and biases are reinitialized, and the network is trained from scratch using the adversarially augmented training set. A flowchart depicting this augmentation approach is shown in Figure \ref{fig:AugmentationFlowchart}.

\begin{figure}[!htb]
     \centering
     \includegraphics[width=0.49\textwidth]{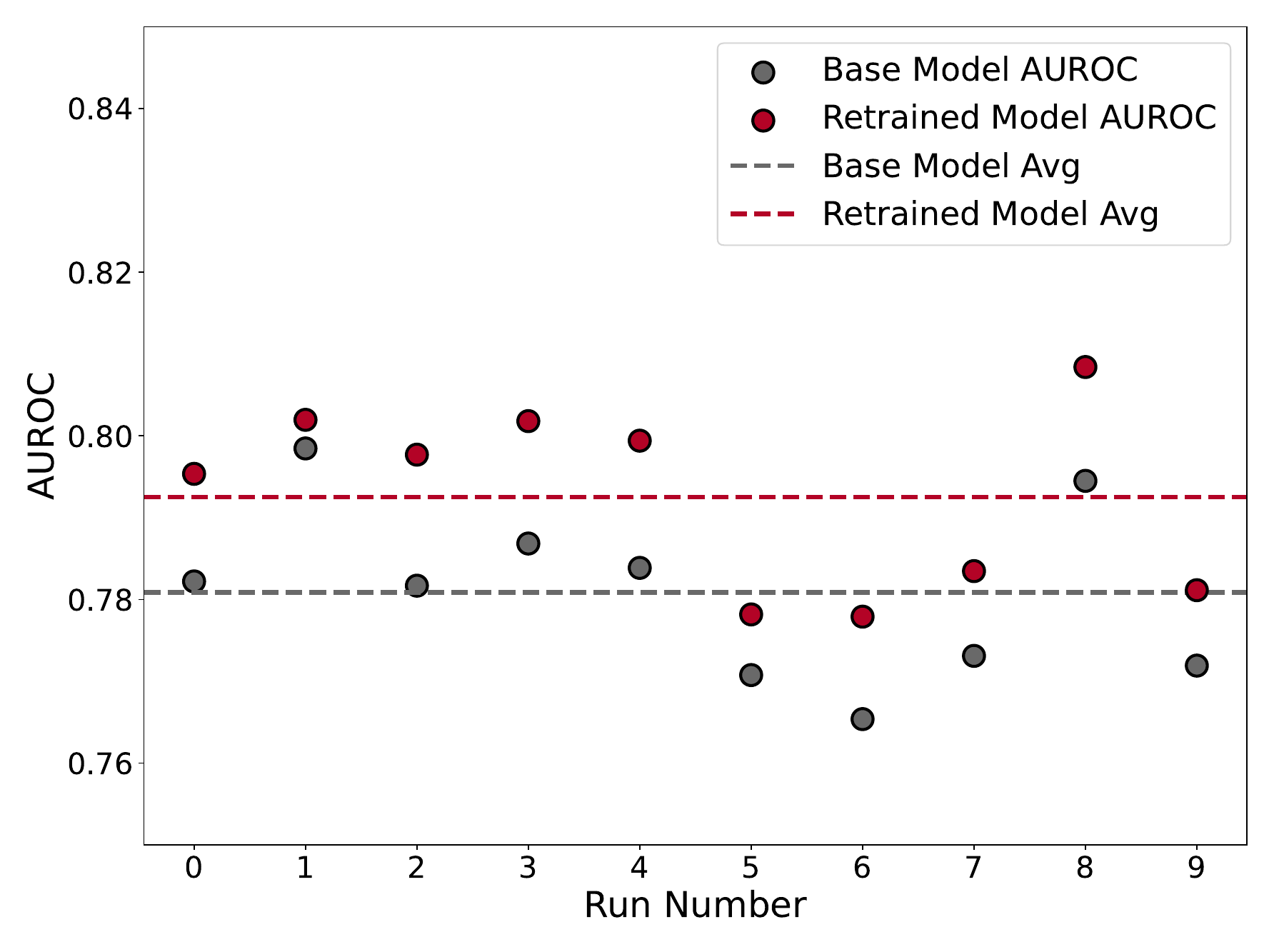}
     \caption{Comparison of the areas under the Receiver-Operating-Curves---on the base model (grey) and the adversarially augmented model (red)---on the Higgs data set over 10 independent runs. The respective average results are shown in dashed lines.}
     \label{fig:Higgs_AUROC_Augmentation}
\end{figure}

To quantify the performance of this augmentation approach, we compare the AUROC on clean test inputs between the baseline network and the network trained on the augmented dataset, averaging the results over 10 independent runs. The corresponding results for the Higgs dataset are shown in Figure \ref{fig:Higgs_AUROC_Augmentation}.

The results show a consistent improvement in performance on clean test data across all runs, demonstrating the effectiveness of using adversarial samples as data-augmentation. Although the average gain is moderate---about one percentage point---it still indicated that adversarial augmentation can enhance generalization---especially in data-limited regimes.

A key limitation for this study is the need to artificially reduce the training set to place the baseline model in the undertrained regime. For the Higgs task especially, achieving this required a significant reduction in training size. However, reducing the dataset too aggressively would negatively impact the adversarial attack itself, which relies on sufficient statistics to optimize perturbations over the full dataset. This invoked a balance between reducing the training set enough to benefit from augmentation, but not so much that the attack itself becomes ineffective.

We expect this trade-off to have a significantly smaller impact for more complex tasks that naturally require much larger training datasets, hence resulting larger performance gains through adversarial augmentation. 

Results for the TT vs. WW dataset are provided in the Appendix \ref{App:AugRes}.

\section{Robustness studies using CONSERVAttack}

A very important field of research within adversarial deep learning---often called adversarial defenses---is studying methods to defend networks against adversarial attacks---e.g. improving a networks adversarial robustness. In this section, we will discuss why it is important in the context of High Energy Physics to conduct such studies, what types of attacks and defenses exist, how they differ and also how they apply to this research. Afterwards, we will apply two well-established defense techniques---adversarial training \cite{goodfellow2015explainingharnessingadversarialexamples} and an adversarial detector \cite{metzen2017detectingadversarialperturbations}---to increase the robustness of the networks for both the Higgs, and the TT vs. WW datasets.

Within security critical fields, such as self-driving cars, it is immediately obvious why it would be important to design networks that are inherently robust to malicious inputs. In HEP however, such security concerns are of very little to no importance. Still, we believe---especially when considering attacks that will not be detected by current sanity checks---it is important to study the robustness of our deep learning models against such cases. Specifically, by testing and improving the networks robustness against invisible attacks, we can get a better understanding of the upper-bound of a models intrinsic uncertainty. 

When it comes to adversarial attacks, they can typically be divided into either two or three classes. White-Box Attacks, Black-Box Attacks, and sometimes even Grey-Box Attacks \cite{akhtar2021advancesadversarialattacksdefenses}. These classes divide the adversaries by the knowledge they have about the model---or the data---they are targeting. In White-Box Attacks, one assumes (nearly) full knowledge about the data and the network, for example knowing the gradients of the loss function of the target network, or knowing the exact training, validation, and test data. In Black-Box Attacks, typically one assumes to have no knowledge about the target network or data. In such cases, the attacks, for example, rely on the capability of querying the target network, hence being able to see whether the model correctly or incorrectly processes the input. Another Black-Box approach is trying to reproduce, as closely as possible, the target network, then generating adversaries on the dummy network, and applying these to the target network. This is often possible due to the phenomenon of transferability of adversarial attacks \cite{papernot2017practicalblackboxattacksmachine}. In Grey-Box Attacks, as the name applies, the information about the network is more nuanced. One assumes to have some knowledge about the network, the data, or the training regime. 

In our case, we deploy a Grey-Box Attack. Here, we have knowledge about the network architecture and the data used for training, validating, and testing the model. We do not however, have any knowledge about the gradients, weights, or biases of the target networks. Initially, we deployed White-Box Attacks, however, it is very hard to construct networks that are truly robust against White-Box Attacks. Additionally, the Grey-Box Attack is more realistic for what we are trying the study---namely under the assumption that such adversaries might also arise from the simulations used in HEP---there we also would not expect the "attack" to have full knowledge about the downstream deep learning network. 

To be precise, our Grey-Box attacks work as follows: First, we use the exact same approach as in the attack studies. We simply apply our adversarial attack algorithm to the test set using the pre-trained model. Doing this, leveraging the gradients of the respective networks, represents a White-Box Attack. However, instead of testing robustness against these attacks, we instead do this process multiple times, each time using the same model architecture and hyperparameters, also fixing the parameters for the adversarial attack---but using random initialization of weights and biases for each of the independent runs. We can then use the adversaries generated on the disjunct runs to test the robustness of the current run against Grey-Box attacks.

To quantify the effectiveness of our adversarial defense techniques, we will compare the fooling ratio achieved on the non-robust base models to the fooling ratio achieved on our---hopefully---more robust models after applying the defense. 

\begin{figure*}[!htb]
     \centering
     \includegraphics[width=0.99\textwidth]{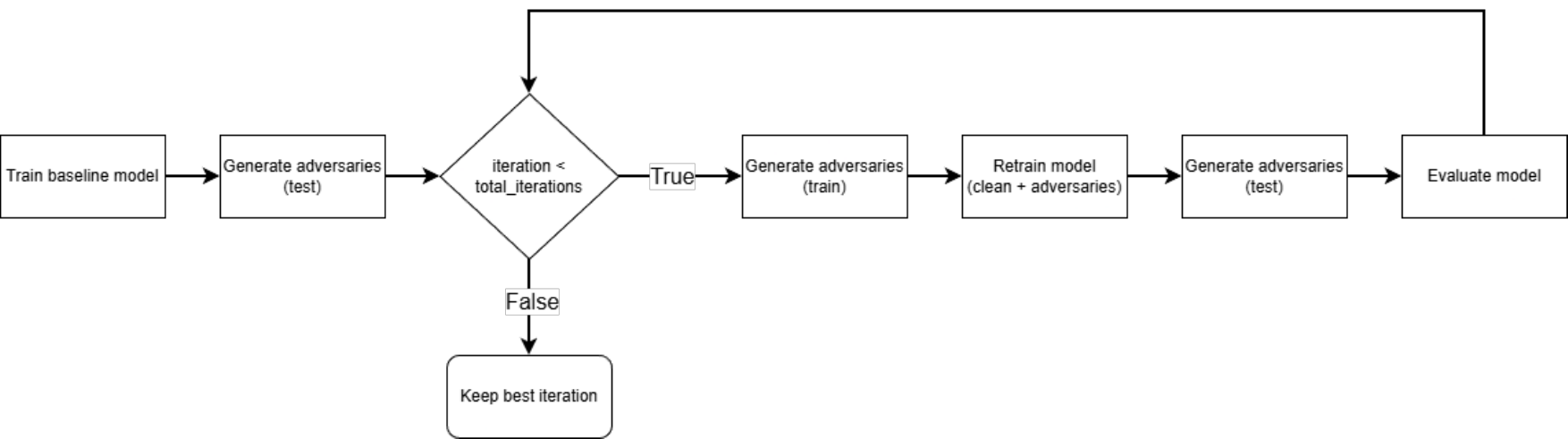}
     \caption{Flowchart showing the workflow used in the adversarial training pipeline.}
     \label{fig:AdversarialTrainingPipeline}
\end{figure*}

For the Adversarial Training approach, we use mostly the same approach as introduced in Section \ref{sec:Augmentation}, however, we train the initial model on the full training set instead of the reduced one. We then generate test adversaries on the initial model. These will later on be used to test the robustness of the adversarially trained models by applying the attacks to each of these models. We then deploy a cumulative iterative adversarial training loop, for the Higgs task, we run 10 iterations, and for the TT vs. WW task we run 5. Within each iteration, we generate train adversarial samples on the current runs initial model using a random initialization of the attack parameters, resulting in different setups for each iteration. After generating these train adversaries, we augment the initial training set cumulatively. This means after the second iteration, we would have a total training set containing the initial training set, the training adversaries generated in iteration 0, and those generated in iteration 1. After each iteration, we save the resulting train adversaries, as well as the resulting adversarially trained models. A flowchart depicting this approach is shown in Figure \ref{fig:AdversarialTrainingPipeline}.

\begin{figure}[!htb]
     \centering
     \includegraphics[width=0.49\textwidth]{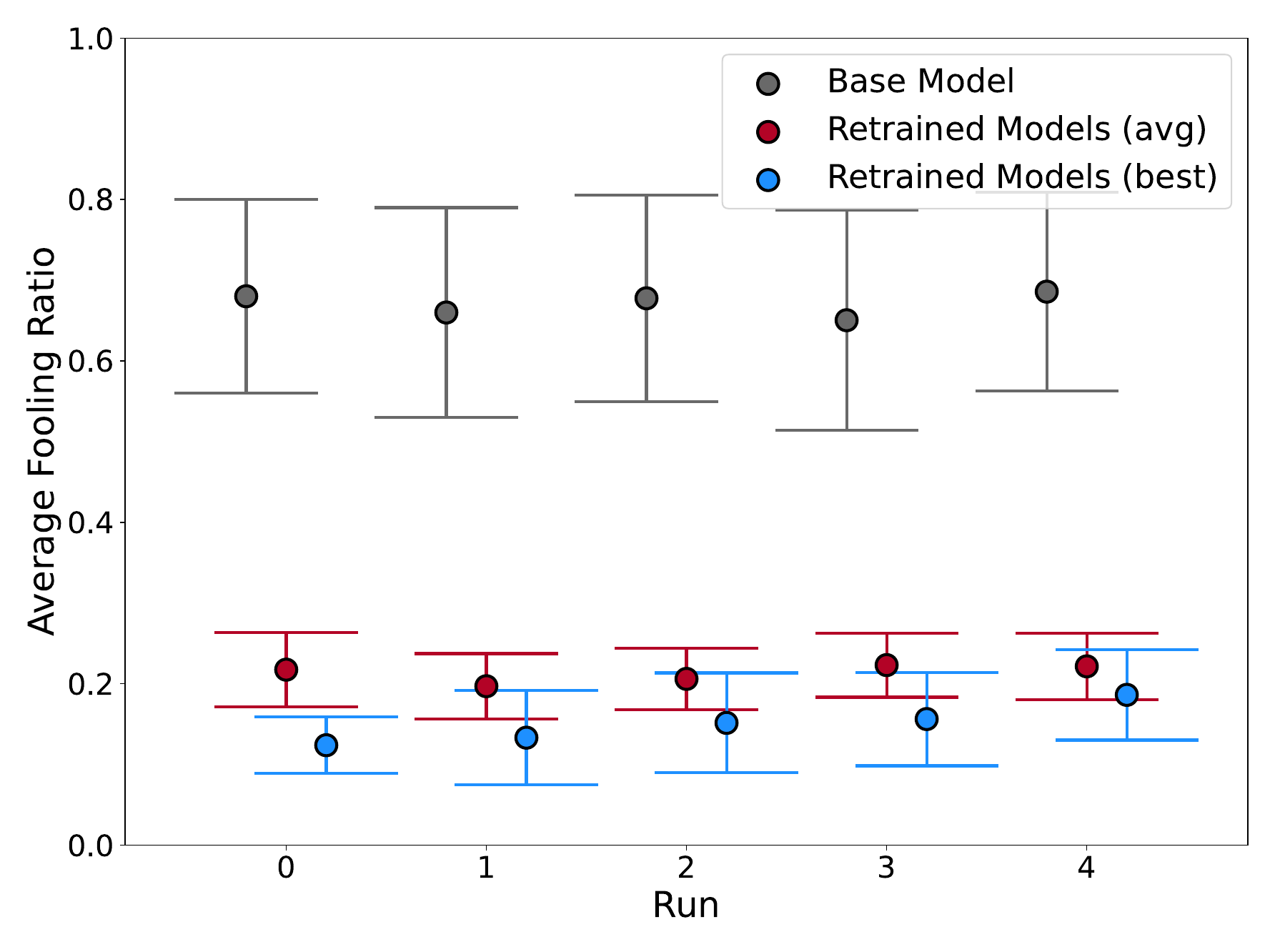}
     \caption{Comparison of the fooling ratios of adversarial attacks on the baseline model (grey) vs the average-case adversarially trained network (red) and the best-case adversarially trained network (blue) on the Higgs dataset.}
     \label{fig:Higgs_AdvTrain_BaseVsRetrained}
\end{figure}

For the Higgs task, we can see the fooling ratio on the base model, as well as the fooling ratio on our cumulatively adversarially trained models in Figure \ref{fig:Higgs_AdvTrain_BaseVsRetrained}.

We see that---on average- the Grey-Box fooling ratio on the base model is around 70\% across all runs. Importantly, here we also include the fooling ratio of the adversaries generated in the same run, which represents a White-Box attack, leading to a slight increase in the average fooling ratio. Additionally, we see that the adversarially trained models exhibit a significantly larger robustness against Grey-Box attacks---both for the average performance across the retraining iterations, as well as choosing the best iteration for each run. We manage to reduce the resulting fooling ratio to about 0.15 when always choosing the best iteration, and to about 0.2 when averaged over all iterations.

\begin{figure*}[!htb]
     \centering
     \includegraphics[width=0.99\textwidth]{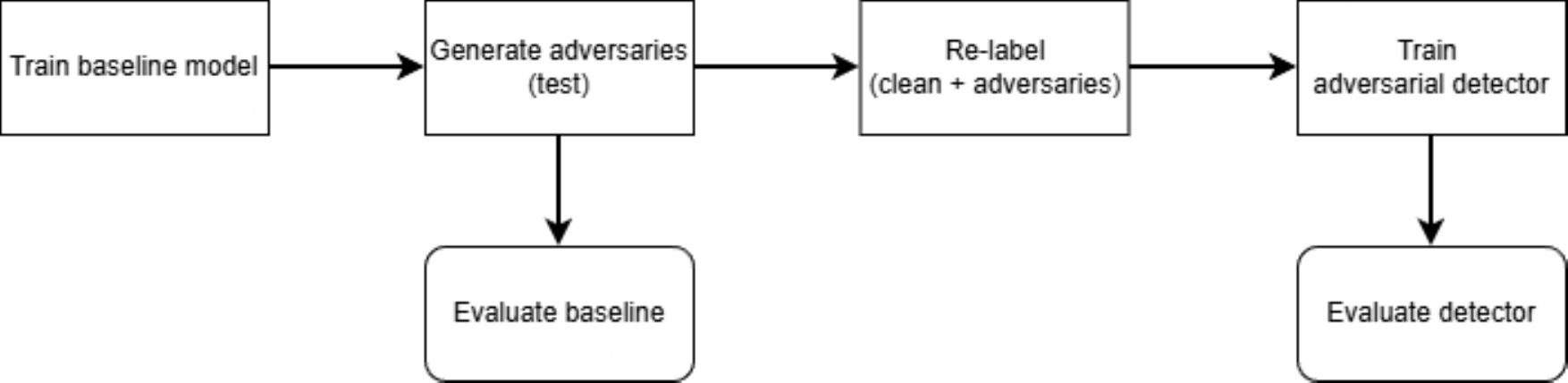}
     \caption{Flowchart showing the workflow used in the adversarial detector pipeline.}
     \label{fig:AdversarialDetectorPipeline}
\end{figure*}

For the Adversarial Detector defense, the pipeline differs slightly compared to the Adversarial Training. While we still train a baseline model with the clean events, instead of generating just adversaries for the train and test sets, here we generate them for all three sets, training, validation, and testing. Then, we re-label both our adversaries, and our clean events, where the adversaries get assigned the label 0, and the clean inputs get 1. 
Afterwards, the Adversarial Detector network is trained to classify between adversaries and clean events. Once trained, we pass the adversarial attacks generated on every other run to the current runs Adversarial Detector. We can then correct the fooling ratio by removing all adversaries that were correctly classified by our Detector. The flowchar for this pipeline is shown in Figure \ref{fig:AdversarialDetectorPipeline}.

\begin{figure}[!htb]
     \centering
     \includegraphics[width=0.45\textwidth]{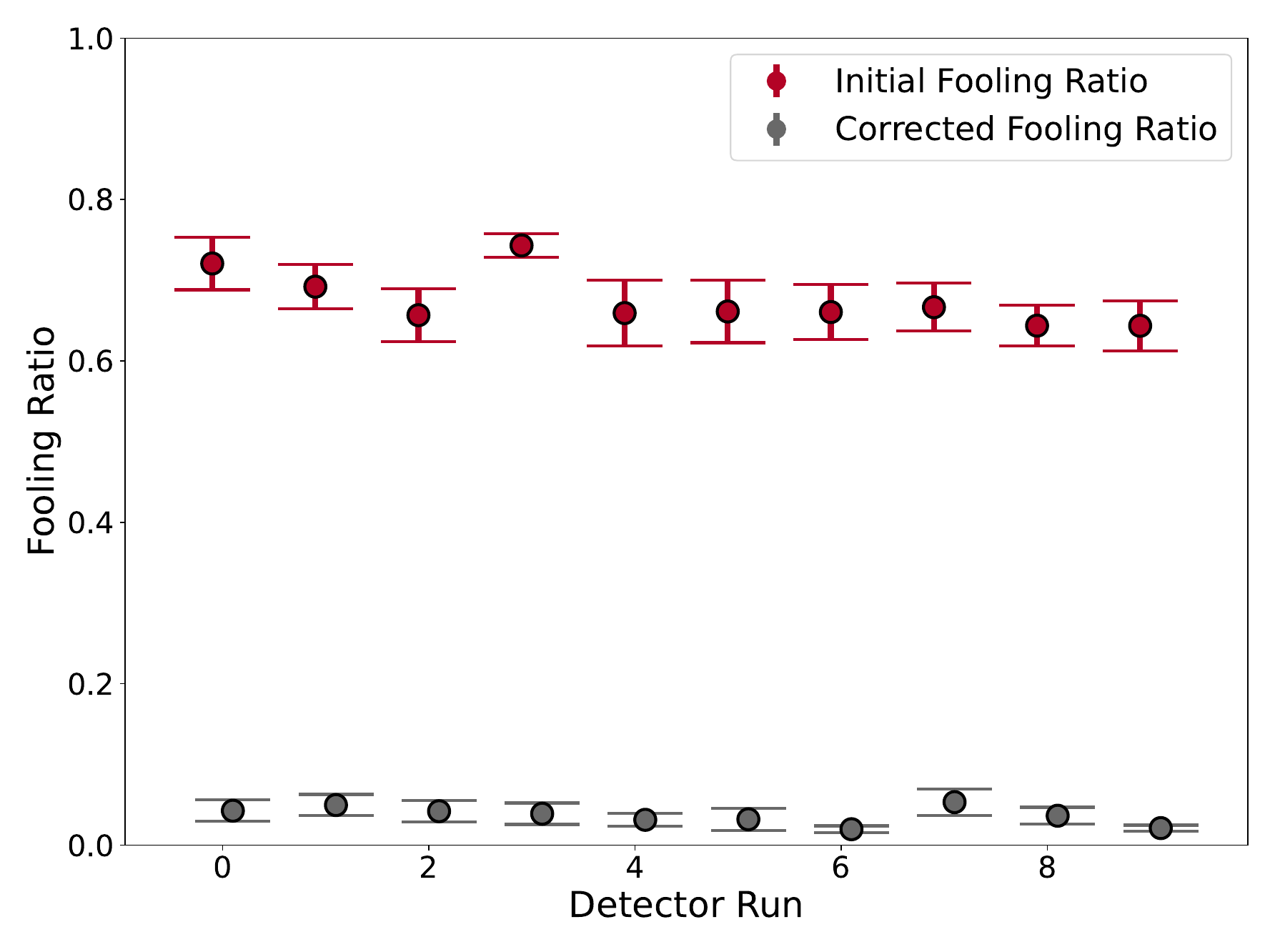}
     \caption{Comparison of the initial fooling ratios (red) and the corrected fooling ratios (grey) found in 10 independent runs of the adversarial detector pipeline on the Higgs dataset.}
     \label{fig:Higgs_AdvDetect_AvgFRScatter}
\end{figure}

For the Higgs dataset, the results can be found in Figure \ref{fig:Higgs_AdvDetect_AvgFRScatter}.

There, we see that the corrected fooling ratio is significantly smaller than the initial. Additionally, this Adversarial Detector increases the robustness even more than Adversarial Training, pushing the fooling ratio down to about 0.05 to 0.08. 

There are, however, distinct trade-offs for both cases. A drawback introduced---briefly introduced previously---about Adversarial Training is, especially in cases where the model has already seen enough data, that such training can decrease the models efficiency on the clean events.

While the same thing can in theory also happen for the Adversarial Detector approach, here it is much more unlikely. A decrease in clean efficiency would only occur if the Adversarial Detector is very bad at identifying clean inputs correctly, which would lead to filtering out a large amount of training data, consequently reducing the networks performance. However, in the studies here, the detectors were always significantly better than chance guessing at classifying between clean and adversarial events, leading to a negligible reduction in training size.

In both defenses, generating the required adversaries, as well as the training and fine-tuning of the respective models introduces computational overhead, therefore increasing the workload during the production pipeline. The Adversarial Detector also increases inference time, since an additional model query is required: inputs must first be passed through the Detector to filter out events classified as adversarial.

However, applying these techniques provides a better understanding of the worst-case systematic uncertainties of the underlying deep learning models.

\section{Adversarial Detector: Systematic Misclassifications and Real-Data Evaluation}

In addition to the robustness studies conducted on the Adversarial Detector discussed in the previous section, we further examine two aspects of the Adversarial Detector's behavior.

First, we would investigate whether the clean events that are misclassified as adversaries represent systematic effects or whether they can be attributed to statistical fluctuations within the network. To this end, we model the misclassification of clean events across multiple runs as independent Bernoulli trials. Let each event be evaluated over \(R\) independent runs. For run \(i\), we estimate the empirical detector accuracy

\begin{align}
\text{acc}_i := \frac{N_{\text{correct}} - N_{\text{flagged}}}{N_{\text{correct}}},
\end{align}

where $N_{\text{correct}}$ is the number of correctly classified samples and
$N_{\text{flagged}}$ is the number of detector-flagged misclassifications.
Then, the average empirical accuracy over all runs is

\begin{align}
\bar{a} := \frac{1}{R} \sum_{i=1}^{R} \text{acc}_i,
\end{align}

and the per-run misclassification probability is defined as
\begin{align}
p := 1 - \bar{a}.
\end{align}

Under the null hypothesis that misclassification events occur independently for each sample and run with probability \(p\), the number of misclassifications for an event follows a binomial distribution. The probability that an event is misclassified in exactly \(k\) of the \(R\) 
runs is
\begin{align}
P(K = k) 
:= \binom{R}{k} p^{k} (1-p)^{R-k}.
\end{align}

Given \(N\) total events, the expected number of events misclassified exactly \(k\) times is
\begin{align}
E_k := N \cdot P(K = k).
\end{align}

Let \(O_k\) denote the observed number of events that were misclassified exactly \(k\) out of \(R\) runs. For bins in which \(E_k\) is small, we approximate the binomial fluctuation by a Poisson distribution with rate parameter \(\lambda = E_k\). The significance of observing at least \(O_k\) samples in bin \(k\) is assessed using the Poisson upper-tail probability

\begin{align}
p_{\text{val}}(k)
&:= P\!\left( X \ge O_k \,\middle|\, X \sim \mathrm{Poisson}(E_k) \right) \nonumber\\
&= \sum_{x = O_k}^{\infty} \frac{E_k^{\,x} e^{-E_k}}{x!}.
\end{align}

\begin{figure}[!htb]
     \centering
     \includegraphics[width=0.49\textwidth]{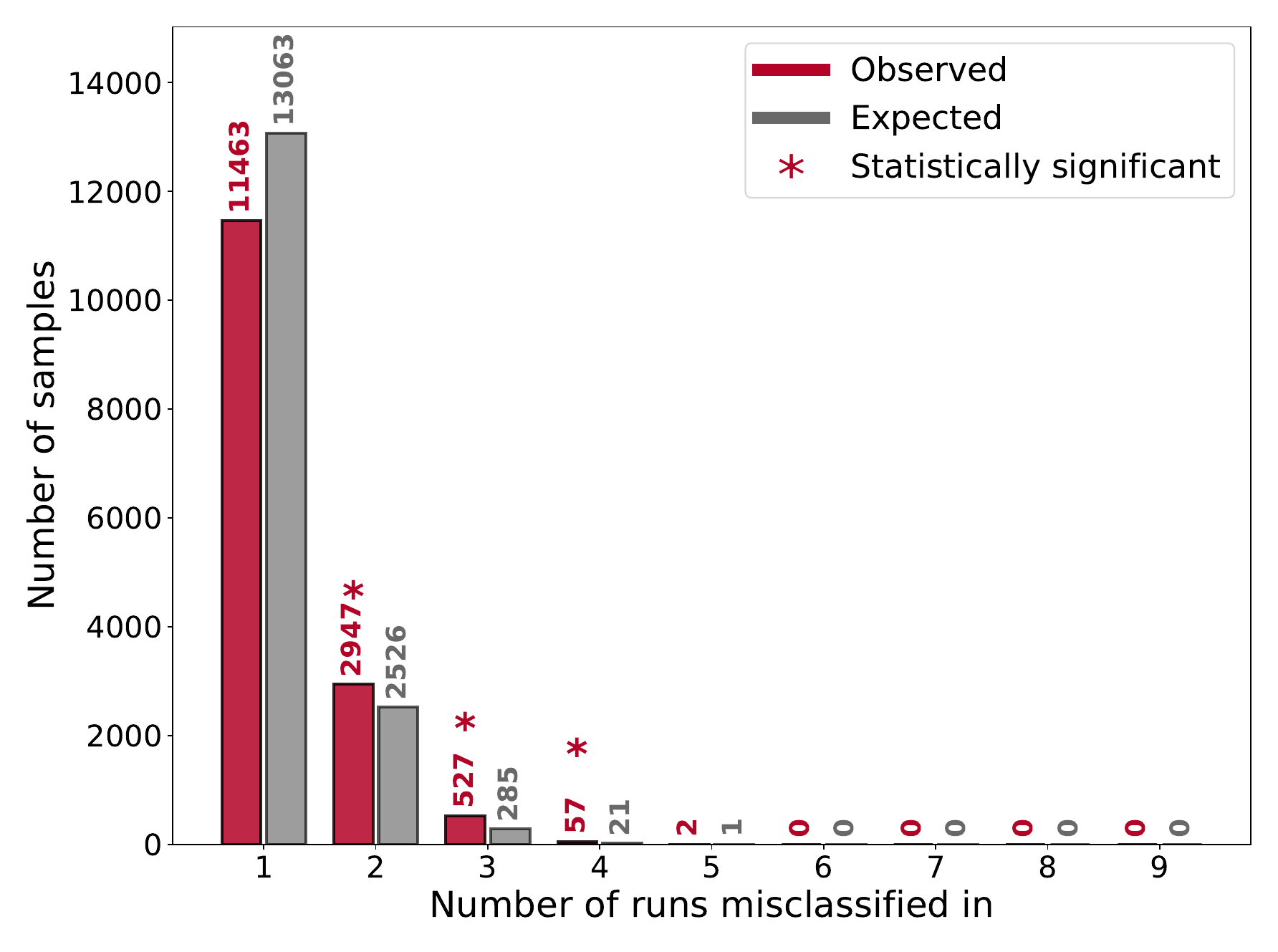}
     \caption{Comparing the expected amounts of events misclassified across multiple runs (grey) to the observed amount (red). The red stars depict statistically significant differences between observed and expected. This plot was generated using the TT vs. WW dataset.}
     \label{fig:Topo_AdvDetect_MisclassBins}
\end{figure}

Bins with \( p_{\text{val}}(k) < 0.05 \) are considered to show a statistically significant excess of repeated misclassifications, indicating that there are indeed some clean events exhibiting adversarial-like behavior and are consistently misidentified by 
the detector beyond what would be expected from random fluctuations. The corresponding plot obtained when applying this procedure to the TT vs. WW data is shown in Figure \ref{fig:Topo_AdvDetect_MisclassBins}.

There, we observe that the number of clean inputs misclassified repeatedly across multiple runs---specifically in two, three, and four out of the ten runs---exceeds the expectation under the binomial-statistical model described above at a significant level. Consequently, indicating that a subset of clean events systematically appears adversarial to the detector, rather than these repeated misclassifications being attributable merely to random fluctuations during training. In other words, the detector consistently struggles with certain clean events in a reproducible way, suggesting that they might share structural properties with genuine adversaries as perceived by the learned representation.
\newline
\newline
Next, we evaluate the performance of the adversarial detector on real data rather than on simulated events. For this study, we focus on the TT vs. WW task and use real collision data from the 2012 CMS Single Mu dataset \cite{SingleMu}. These events are processes in the same way as the simulated samples, and are labeled as "clean" inputs for evaluation in the adversarial detector. Importantly, none of these real events were used during the training of the detector network.

\begin{figure}[!htb]
     \centering
     \includegraphics[width=0.49\textwidth]{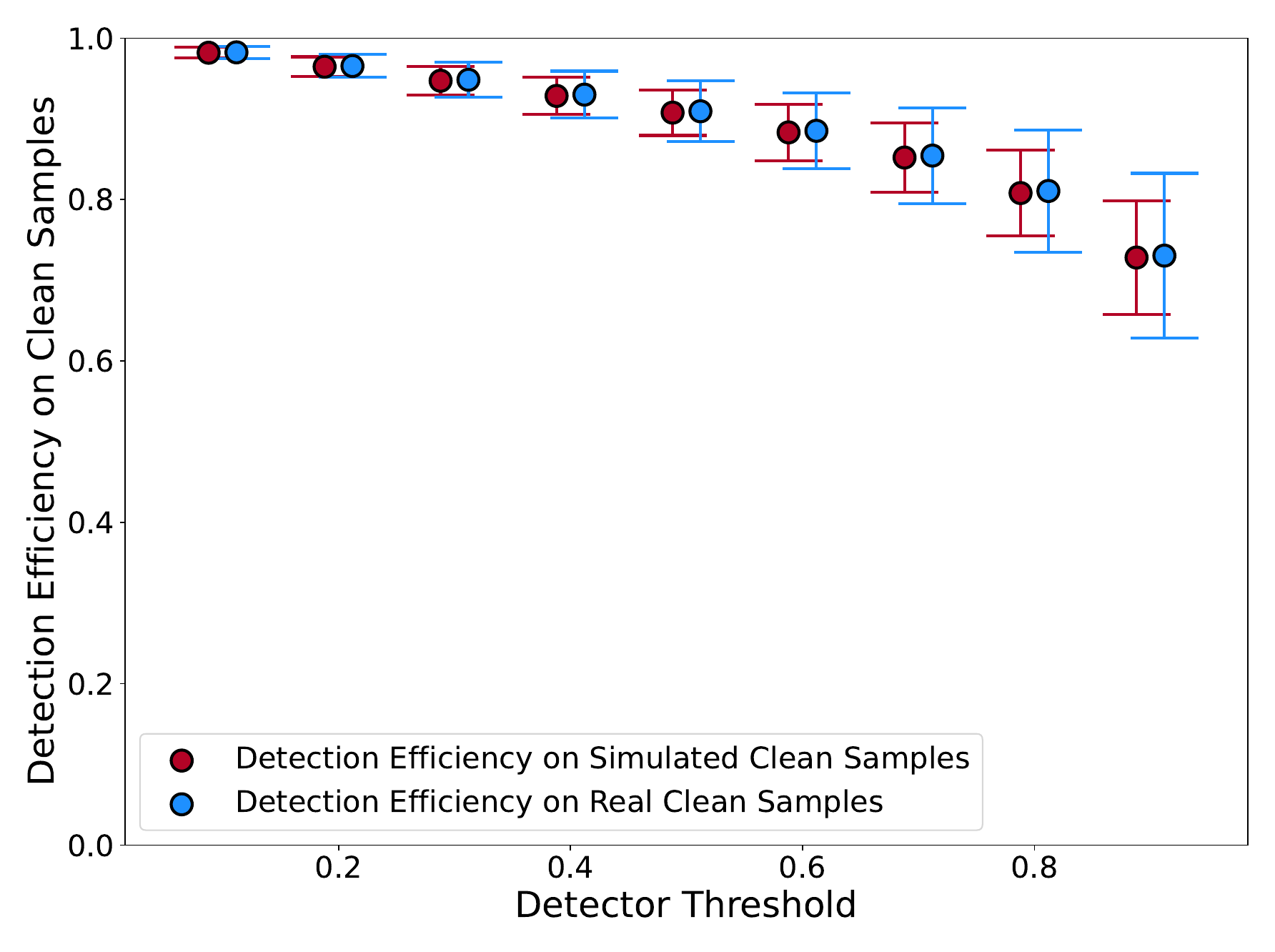}
     \caption{Detector accuracies on both clean non-simulated clean events (blue) and simulated clean events(red) using an adversarial detector network trained solely on clean and adversarial simulated events, dependent on the classification threshold for the TT vs. WW dataset.}
     \label{fig:Topo_AdvDetect_CleanSimRealEfficiency}
\end{figure}

To assess the detector's behavior under domain shift, we measure its clean-sample detection efficiency---the fraction of clean events classified as non-adversarial---for both real and simulated events across varying detector confidence thresholds, averaging the results over all detector runs. The corresponding comparison is shown in Figure \ref{fig:Topo_AdvDetect_CleanSimRealEfficiency}.

The adversarial detector manages to maintain a high confidence in correctly classifying clean events; even at high decision thresholds (greater than 0.8), the clean-event efficiency remains around 0.8. Notably, the detector performs almost equally well on real events, despite never having encountered them during training. While the mean detection efficiency is nearly identical to that of the simulated events, particularly at higher thresholds, the run-to-run variance increases, indicating a reduction in overall stability for real events.

These results suggest that there is no substantial domain gap between simulated and real clean events---at least not one that causes the detector to misidentify real events as adversaries. Consequently, the well-known degradation in model performance when applying models trained on simulated events to real data likely cannot be explained solely by real events acting as adversaries to the classifier, or vice-versa as simulated events poisoning the networks. Nonetheless, the detector does misclassify a fraction of real clean events as adversarial. Therefore, one cannot assume with certainty that all real clean events behave like their simulated counterparts, nor that simulated clean events are never adversarial relative to the model evaluated on real data.

These observation also suggest that the adversarial detector leverages higher-dimensional structure in the datasets, rather than solely on marginal distributions or correlations. Although the adversarial examples are explicitly constructed to remain indistinguishable from clean events under standard statistical comparisons, the detector still identifies them as anomalous. This implies that the model is sensitive to complex, non-linear features relationships that are not captured by traditional HEP sanity checks, and that clean events---whether real or simulated---occupy a region of the feature space that remains well separated from the adversarial events.

\section{Donut: Illustrative Example}\label{sec:donut}
In this section, we provide an illustrative example demonstrating how the proposed algorithm constructs adversarial events. Furthermore, as previously highlighted, an adversarial detector---a deep learning network trained to classify between clean and adversarial events---represents a promising approach to improving a network’s robustness against adversarial attacks. We therefore further use this example to gain insight into how the adversarial detector operates.

To this end, we construct a toy dataset consisting of two input variables $x_1$ and $x_2$, and two target classes: signal and background. Signal events are sampled from a two-dimensional Gaussian distribution centered at the origin, where each coordinate $(x_1, x_2)$ is drawn independently from a normal distribution with mean $0$ and standard deviation $\sigma$, i.e., $x_1, x_2 \sim \mathcal{N}(0,\,\sigma^2)$. 

Background events are sampled from a donut-shaped distribution, where the radius $r$ is drawn from a normal distribution centered at $r_\text{ring}$ with standard deviation $\sigma$, $r \sim \mathcal{N}(r_\text{ring},\,\sigma^2)$, and the angle $\theta$ is drawn uniformly from $[0, 2\pi)$, $\theta \sim \mathcal{U}(0, 2\pi)$. The corresponding Cartesian coordinates are given by $x_1 = r \cos\theta$ and $x_2 = r \sin\theta$. For both classes, the input variables are linearly uncorrelated.

\begin{figure}[!htb]
     \centering
     \includegraphics[width=0.49\textwidth]{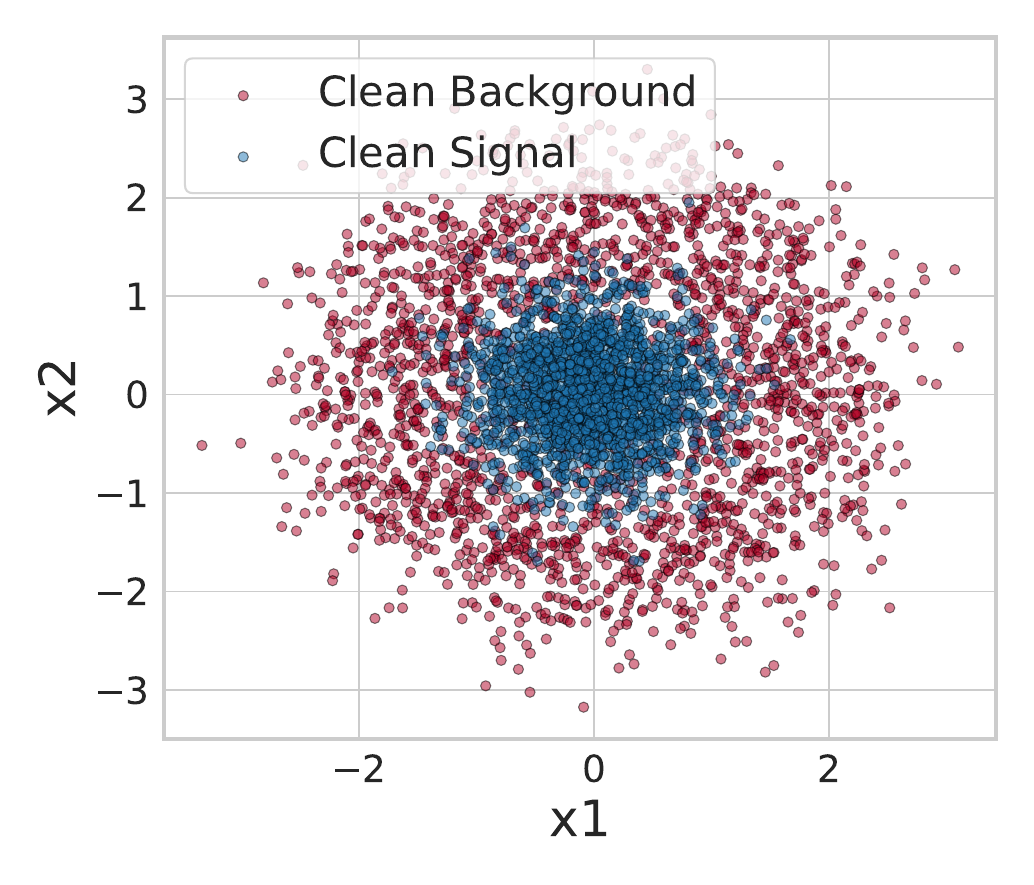}
     \caption{Point clouds of sub-samples points from both the clean signal (blue) and clean background (red) events for the Donut toy dataset.}
     \label{fig:Donut_Signal_BG_Pointcloud}
\end{figure}

The resulting two-dimensional distributions are shown in Figure \ref{fig:Donut_Signal_BG_Pointcloud}.

The signal distribution forms a circle in the center of the features space, which is surrounded by the donut-shaped background distribution, with a small overlap between the two classes.

To visualize both the attack mechanism and the behavior of the adversarial detector, we apply the same studies discussed in previous sections to this toy dataset. We first generate adversaries on the training and test sets, which are subsequently used to train the adversarial detector network. Here, adversaries are generated exclusively on background events. 

Using this procedure, we achieve an initial fooling ratio of 0.354 on the signal-background classifier. After applying the adversarial detector, this value is reduced to a corrected fooling ratio of 0.153, corresponding to a detector efficiency of 0.568 on adversarial events. For clean events, the detector achieves a higher efficiency of 0.788.

Compared to the results presented earlier, both the initial fooling ratio and the detector efficiency on adversarial events are relatively low. We attribute this to the low dimensionality of the dataset: since the data are described by only two input variables, the attack can perturb only these same two features in order to induce misclassification. At the same time, the attack remains constrained to minimally alter the marginal distributions and correlations, making successful misclassification substantially more difficult and thereby limiting the achievable fooling ratio.

A similar limitation applies to the adversarial detector. With only two input variables, the decision boundary that separates clean and adversarial events must also be defined solely in this two-dimensional space, making effective separation challenging. As illustrated in the following, many adversarial events lie well within the signal region, such that correctly identifying them as adversaries would require misclassifying a significant fraction of clean signal events.

To illustrate how the attack modifies the underlying data, we present two complementary visualizations. First, we compare the one-dimensional marginal distributions of the clean background events to those of the adversarial background events. Second, we visualize the corresponding two-dimensional feature distributions as point clouds, allowing for a direct inspection of how the adversarial perturbations reshape the data in feature space.

\begin{figure}[!htb]
     \centering
     \includegraphics[width=0.49\textwidth]{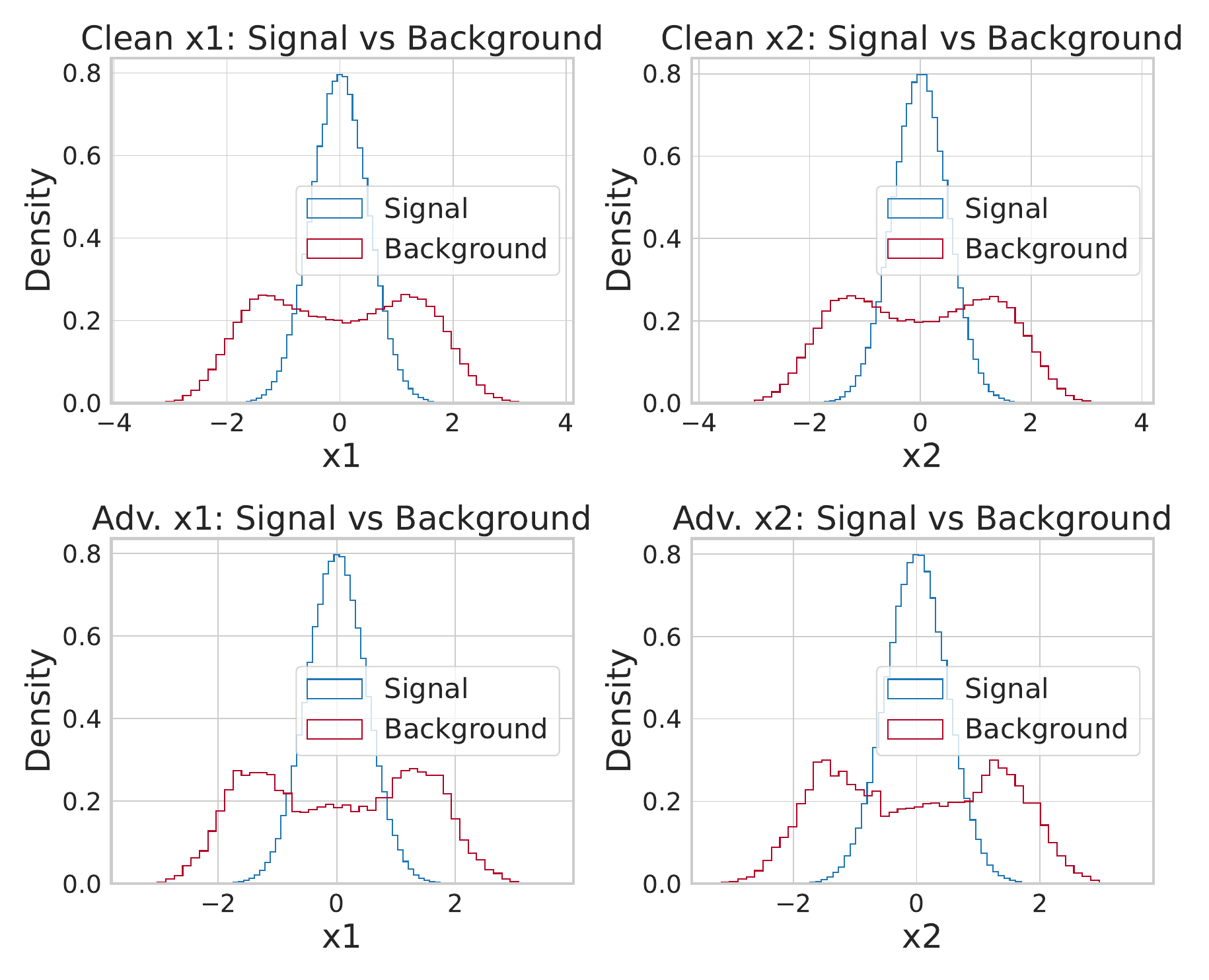}
     \caption{Comparison of the feature distributions between clean signal vs. clean background events (top row) and clean signal vs. adversarial background events (bottom row).}
     \label{fig:Donut_Dists_CleanAdv}
\end{figure}

\begin{figure}[!htb]
     \centering
     \includegraphics[width=0.49\textwidth]{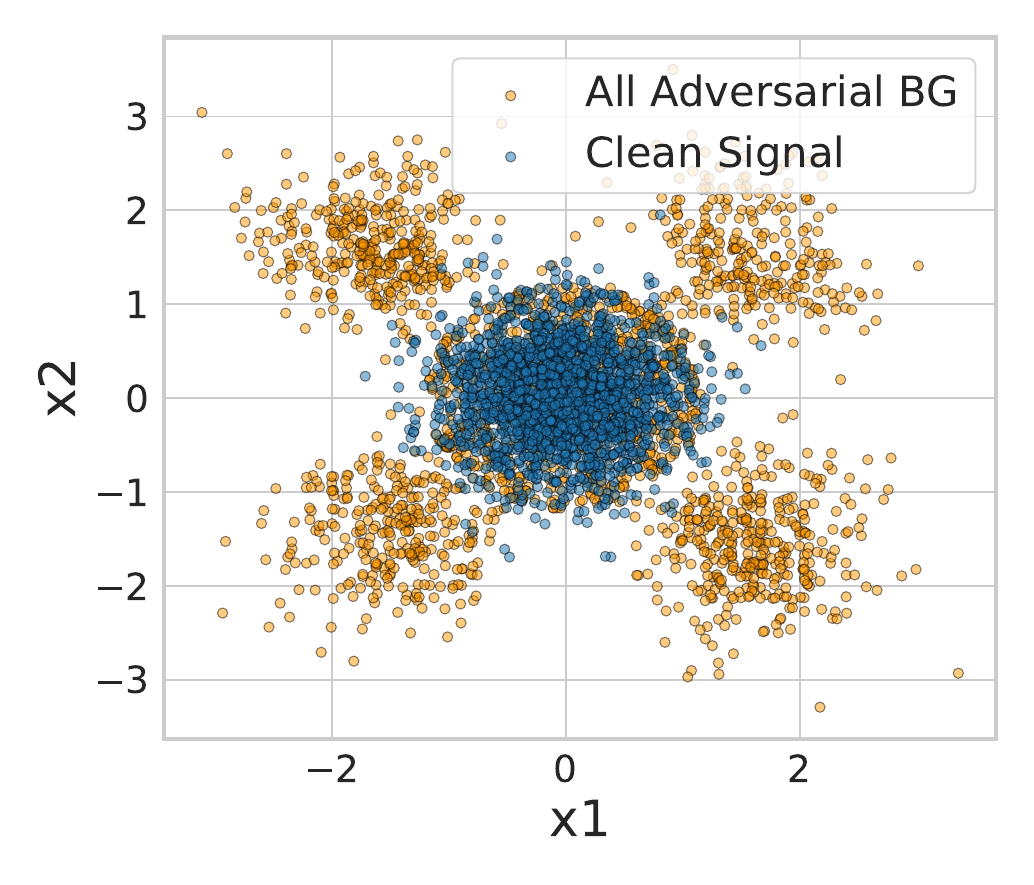}
     \caption{Point clouds of sub-samples points from both the clean signal (blue) and adversarial background (yellow) events for the Donut toy dataset.}
     \label{fig:Donut_Pointcloud_AdvSignal}
\end{figure}

The resulting marginal distributions are shown in Figure \ref{fig:Donut_Dists_CleanAdv}. For both clean and adversarial background events, the marginals differ noticeably from those of the signal, as expected: the signal is distributed according to a two-dimensional Gaussian centered at the origin, leading to pronounced peaks at zero in both features, whereas the donut-shaped background has a depletion in the central region, resulting in a characteristic dip in the one-dimensional projections.

In this example, the differences between clean and adversarial background distributions are quite prominent. This can likely be attributed to the low dimensionality of the dataset: since the attack can only perturb two features, achieving misclassification requires comparatively larger modifications to each individual variable, as the perturbations cannot be distributed across a higher-dimensional feature space. Nevertheless, the adversarial background marginals preserve the qualitative structure of the original background distribution, maintaining a central depletion surrounded by two smaller peaks.

The two-dimensional point cloud comparing adversarial background events to clean signal events is shown in Figure \ref{fig:Donut_Pointcloud_AdvSignal}. In this representation, the effect of the attack becomes substantially more apparent than in the one-dimensional marginals. A large fraction of background events is shifted into the central region of feature space that is populated by signal events, effectively overlapping with the signal distribution around the origin.

\begin{figure}[!htb]
     \centering
     \includegraphics[width=0.49\textwidth]{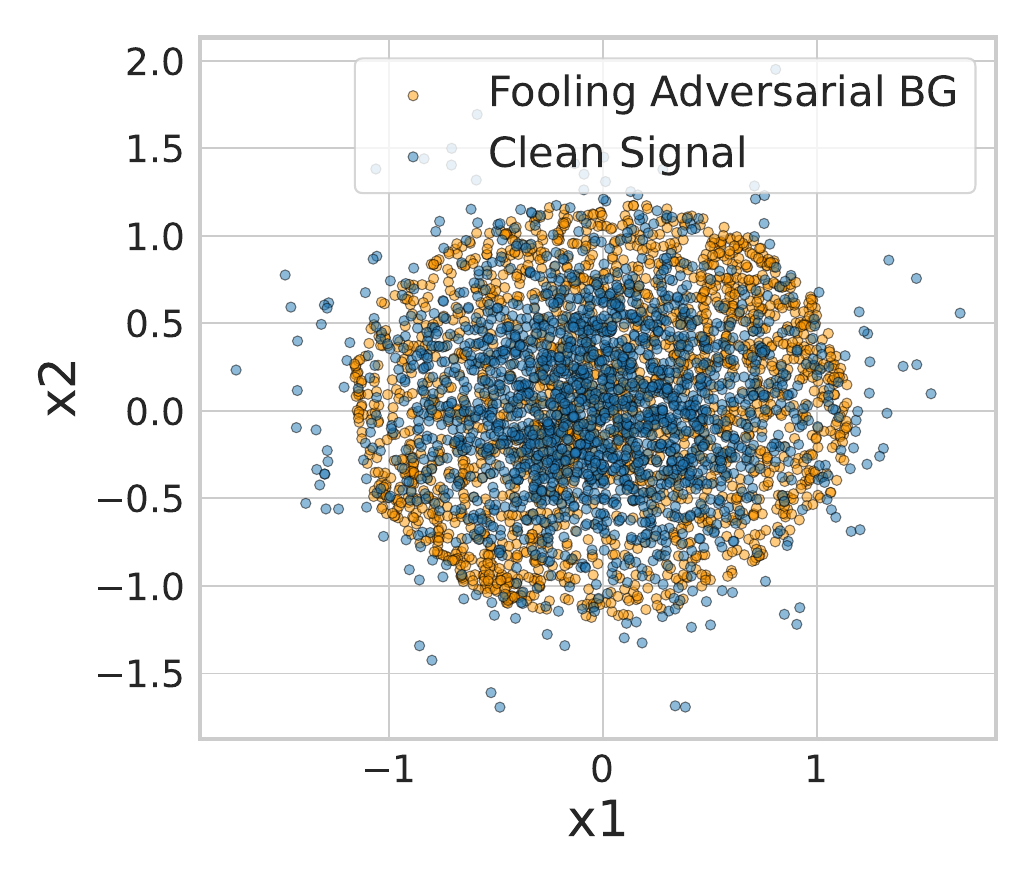}
     \caption{Point clouds of sub-samples points from both the clean signal (blue) and adversarial background events that successfully fool the baseline network (yellow)  for the Donut toy dataset.}
     \label{fig:Donut_Pointcloud_FoolingAdvSignal}
\end{figure}

Notably, many of the background events that are displaced into the signal region originate from areas along the coordinate axes, as evidenced by the depletion of events along these axes outside the central region after the attack. Importantly, however, not all adversarial background events lead to a misclassification by the original classifier. Restricting the visualization to only those adversarial events that successfully fool the network yields the point cloud shown in Figure \ref{fig:Donut_Pointcloud_FoolingAdvSignal}, where, as expected, the fooling adversaries are concentrated almost exclusively within the central signal region.

\begin{figure}[!htb]
     \centering
     \includegraphics[width=0.49\textwidth]{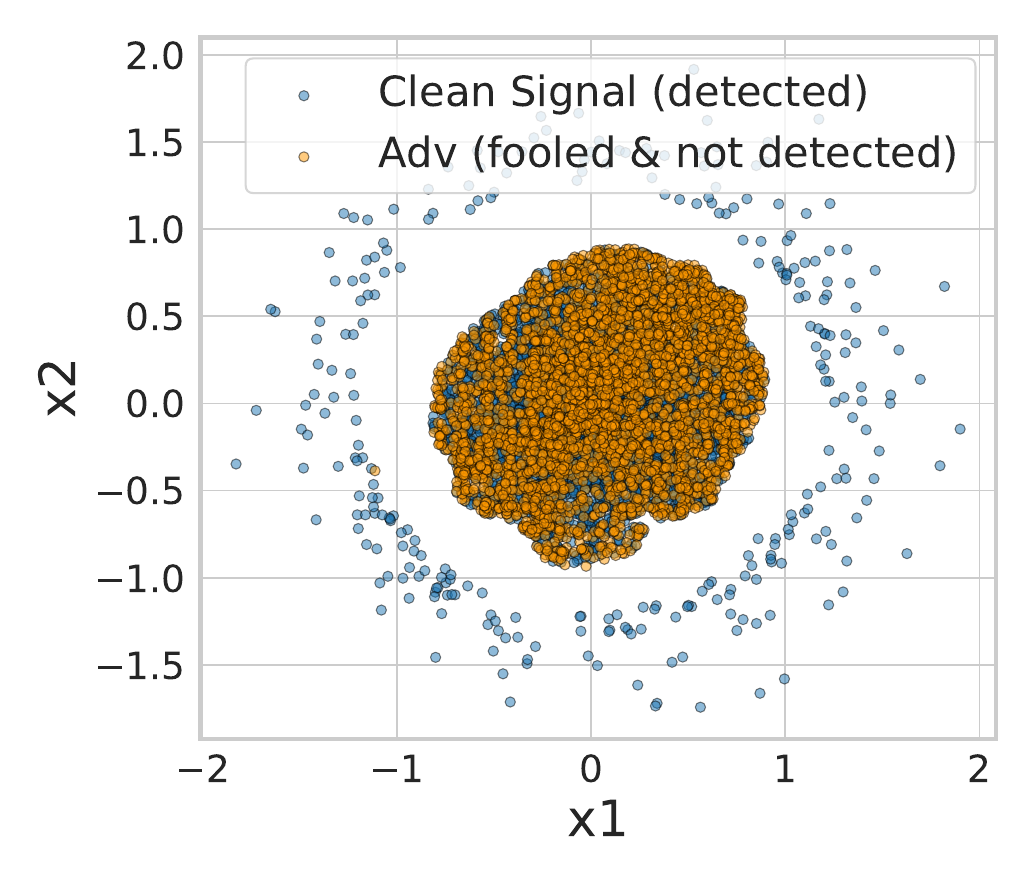}
     \caption{Point clouds of sub-samples points from both the clean signal correctly classified by the adversarial detector network as such (blue), and adversarial background events that successfully fool the baseline network but are not correctly identified as such by the detector network (yellow)  for the Donut toy dataset.}
     \label{fig:Donut_Pointcloud_DetectedAdvSignal}
\end{figure}

We now train an adversarial detector network to classify between adversarial background events and clean events, including both signal and background. Applying this detector to the clean signal events and the adversarial background events in the test set yields the point cloud shown in Figure \ref{fig:Donut_Pointcloud_DetectedAdvSignal}.

This point cloud shows clean signal events that are correctly identified as such by the detector, together with adversarial background events that fool the original classifier---i.e., are classified as signal by the initial network---but are not recognized as being adversarial by the detector network. The visualization highlights the strategy employed by the adversarial detector: it learns a decision boundary that balances the correct classification of clean and adversarial events while minimizing the overall misclassification rate. While this approach is standard in deep learning, such decision boundaries are seldom directly visualizable in practice, as most datasets are very high-dimensional.

In this section, we leveraged a simple two-dimensional dataset with clearly separated signal and background classes to illustrate both the mechanism of the adversarial attack and the operation of the adversarial detector network. Although the datasets low dimensionality naturally limits the effectiveness of the attack and negatively impacts the detector’s performance, it provides an opportunity to visualize the underlying concepts in a clear and intuitive way. By examining these toy distributions, we gain concrete insight into how adversarial perturbations shift events in feature space and how the detector learns to differentiate between clean and adversarial events, insights that are often difficult to extract from high-dimensional datasets.

\section{Distance Correlation Studies}

In the previous studies, we focused on preserving linear correlations between features by minimizing changes to the Pearson correlation matrices of clean and adversarially perturbed datasets. While this approach is well aligned with standard validation and sanity checks in high energy physics, it captures only linear dependencies and therefore does not fully reflect the complete statistical structure of the data. Non-linear correlations can also encode important information, both in physics and in other scientific domains.

To address this limitation, we extend the adversarial attack to preserve non-linear dependencies by replacing the Pearson correlation constraint with Distance Correlation \cite{Sz_kely_2007}. We then study this modified adversarial construction within the same adversarial detector framework used in the previous sections, applying it to the Higgs dataset. The hyperparameter choices for the adversarial attack are summarized in the Appendix \ref{App:AttackParams}.

Distance correlation is a statistical measure that detects any form of dependence between random variables, linear or non-linear. For two random vectors $X \in \mathbb{R}^p$ and $Y \in \mathbb{R}^q$, the distance correlation $\mathcal{R}(X, Y)$ is zero if and only if $X$ and $Y$ are independent. It is computed by first forming pairwise Euclidean distance matrices for $X$ and $Y$, double-centering these matrices (subtracting row and column means and adding the overall mean), and then calculating the covariance between the centered distances, normalized by the respective distance variances. Optimizing adversarial perturbations with respect to distance correlation ensures that both linear and nonlinear feature dependencies are preserved, maintaining the data’s statistical structure.

In addition to being more precise at capturing both linear and non-linear feature dependencies, using the Distance Correlation also imposes an additional restrictive burden on the attack algorithm, making it harder to find adversaries compared to the Pearson correlation. Moreover, the computation of distance correlation itself is substantially more expensive.

Let $n$ denote the number of events and $d$ the number of features in the dataset. For our perturbations to the Pearson correlation matrix, we do not recompute the full matrix from scratch; instead, we employ an efficient incremental update procedure, described in the Appendix \ref{App:PearsonOptimization}, which exactly updates the means, variances, and covariances. This reduces the per-candidate computational cost from $\mathcal{O}(d^2 n)$ for a full Pearson recomputation to $\mathcal{O}(d)$. In contrast, evaluating a single candidate perturbation using distance correlation requires $\mathcal{O}(d^2 n^2)$ operations due to the need to compute pairwise distances between all events, and to our knowledge no comparable incremental update scheme exists. Consequently, the distance correlation must be fully recomputed for each candidate perturbation, resulting in a substantially higher computational cost.

Due to the substantially increased computational cost, we severely limit the number of adversarial events generated for the distance correlation study. Specifically, we generate a total of 2500 adversarial events for training and 1000 adversarial events for testing. To maintain a balanced training setup, we also restrict the number of clean events used to train the adversarial detector to the same size.

As this represents a rather drastic reduction in available statistics, and in order to isolate the impact of using distance correlation rather than Pearson correlation, we additionally perform a reference study using Pearson correlations with an equally limited number of adversarial events. This allows us to disentangle effects arising from the correlation metric itself from those induced by the reduced sample size. The detailed configuration of the adversarial algorithm for this comparison study is provided in the Appendix \ref{App:AttackParams}.

\begin{figure}[!htb]
     \centering
     \includegraphics[width=0.49\textwidth]{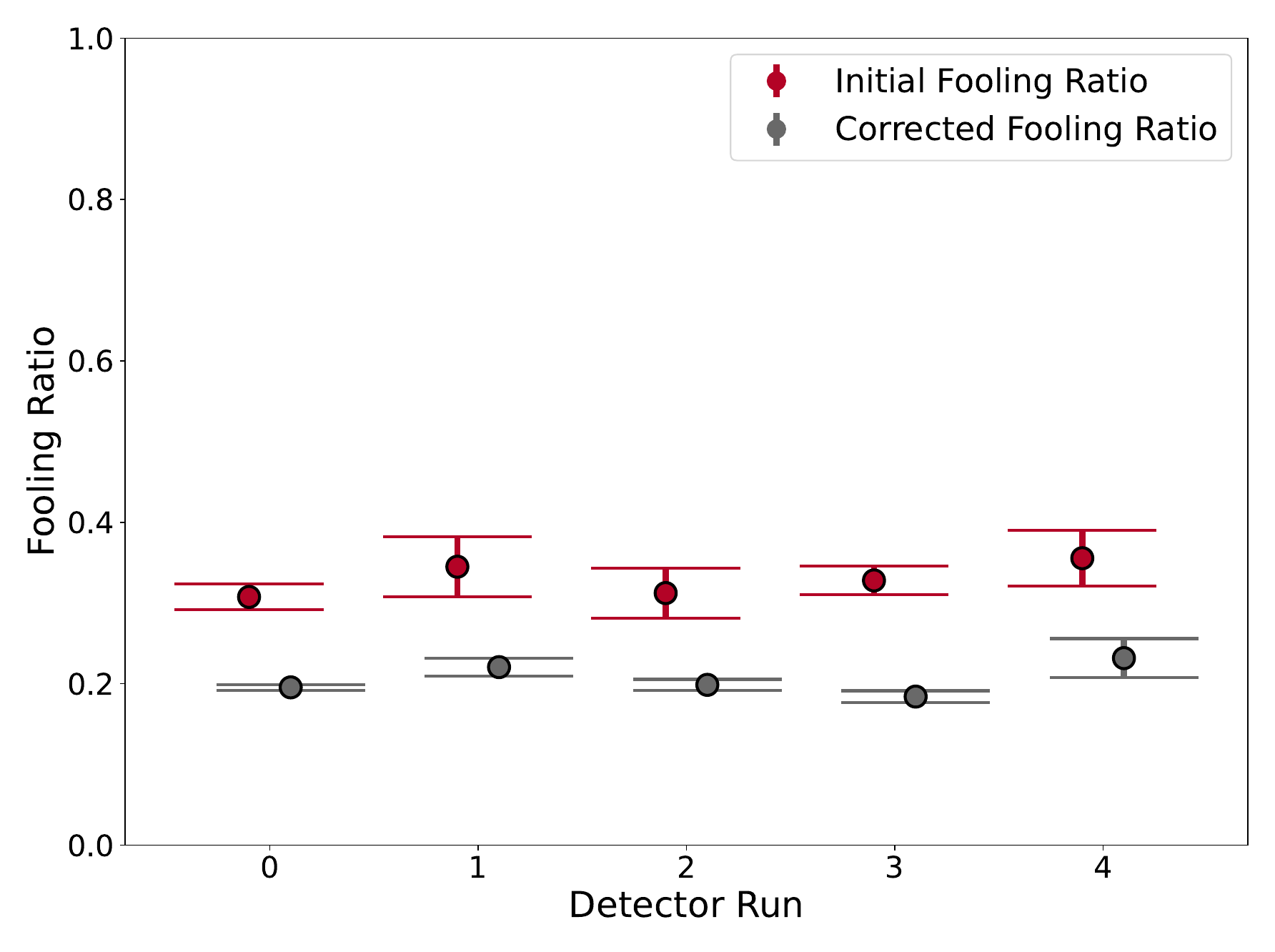}
     \caption{Comparison of the initial fooling ratios (red) and the corrected fooling ratios (grey) found in 5 independent runs of the Distance correlation restricted adversarial detector pipeline on the Higgs dataset.}
     \label{fig:Disco_InitCorrFR}
\end{figure}

The resulting initial fooling ratios, as well as the corrected fooling ratios obtained after applying the adversarial detector pipeline to these adversaries, are shown in Figure \ref{fig:Disco_InitCorrFR}.

\begin{figure}[!htb]
     \centering
     \includegraphics[width=0.49\textwidth]{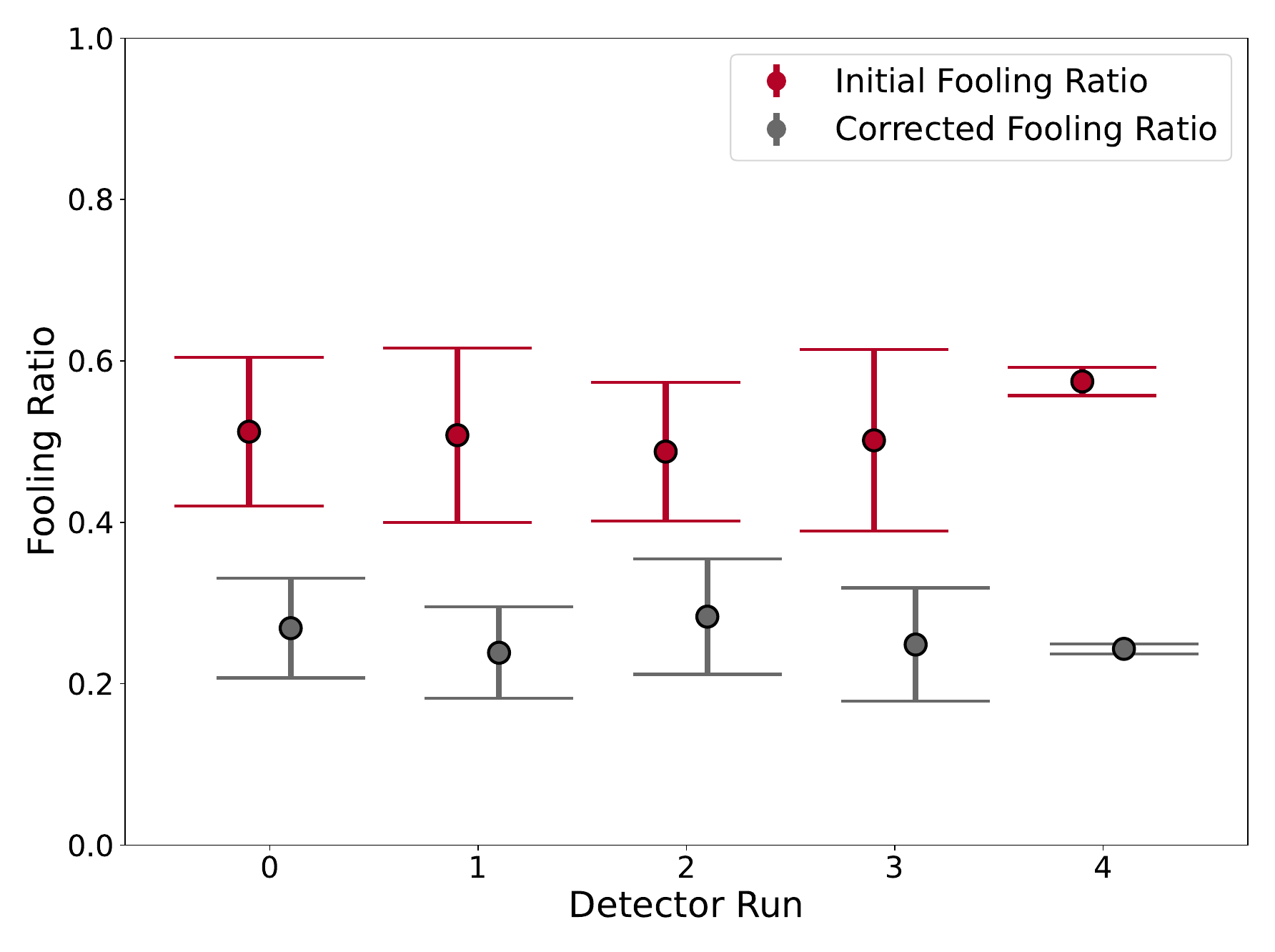}
     \caption{Comparison of the initial fooling ratios (red) and the corrected fooling ratios (grey) found in 5 independent runs of the limited data Pearson correlation restricted adversarial detector pipeline on the Higgs dataset.}
     \label{fig:Disco_LD_InitCorrFR}
\end{figure}

As shown there, the initial fooling ratios are---unsurprisingly---significantly lower than those obtained when generating an unrestricted number of adversaries on the same Higgs dataset using the Pearson correlation. Conversely, the corrected fooling ratios after applying the detector network are considerably higher than in the previous Higgs studies. To gain insight into whether the reduced initial fooling ratio and the increased corrected fooling ratio arise from constraining the adversaries using distance correlation, or instead from the limited number of adversarial samples, we additionally present results from a Pearson-correlation-based attack with the same restricted sample size in Figure \ref{fig:Disco_LD_InitCorrFR}.

From this comparison, we observe that the initial fooling ratios remain significantly lower than those obtained from a Pearson-based attack using a larger adversarial sample size, and that the corrected fooling ratios---consistent with the distance-correlation-based approach---are higher. When directly comparing the limited-sample Pearson and distance-correlation studies, both the initial and corrected fooling ratios are higher for the Pearson case: with the difference being substantial for the initial fooling ratio, while it is comparatively small for the corrected fooling ratio.

Focusing for the moment solely on the initial fooling ratios, the observation that distance-correlation-based adversaries achieve noticeably lower fooling ratios is consistent with the more restrictive nature of this constraint. Put simply, it is substantially harder to construct adversarial examples that satisfy the stricter distance correlation requirements than those based only on Pearson correlations.

We further attribute the reduction in initial fooling ratios observed in the limited-sample regime to the statistically grounded nature of the attack. For larger datasets, perturbing individual events has a comparatively small impact on global statistical properties, such as marginal distributions and both linear and non-linear feature correlations. When the dataset size is reduced significantly, however, each individual perturbation has a much larger influence on these quantities, making it inherently more difficult to generate adversarial events that remain statistically consistent.

An even more pronounced effect is the substantially reduced gap between the initial and corrected fooling ratios when the dataset size is limited, an effect that is observed consistently for both the distance correlation and Pearson correlation studies. While the impact of reduced data size on the initial fooling ratios has already been discussed, we now provide insight into why the corrected fooling ratios are in fact higher than those obtained with larger training samples.

\begin{figure}[!htb]
     \centering
     \includegraphics[width=0.49\textwidth]{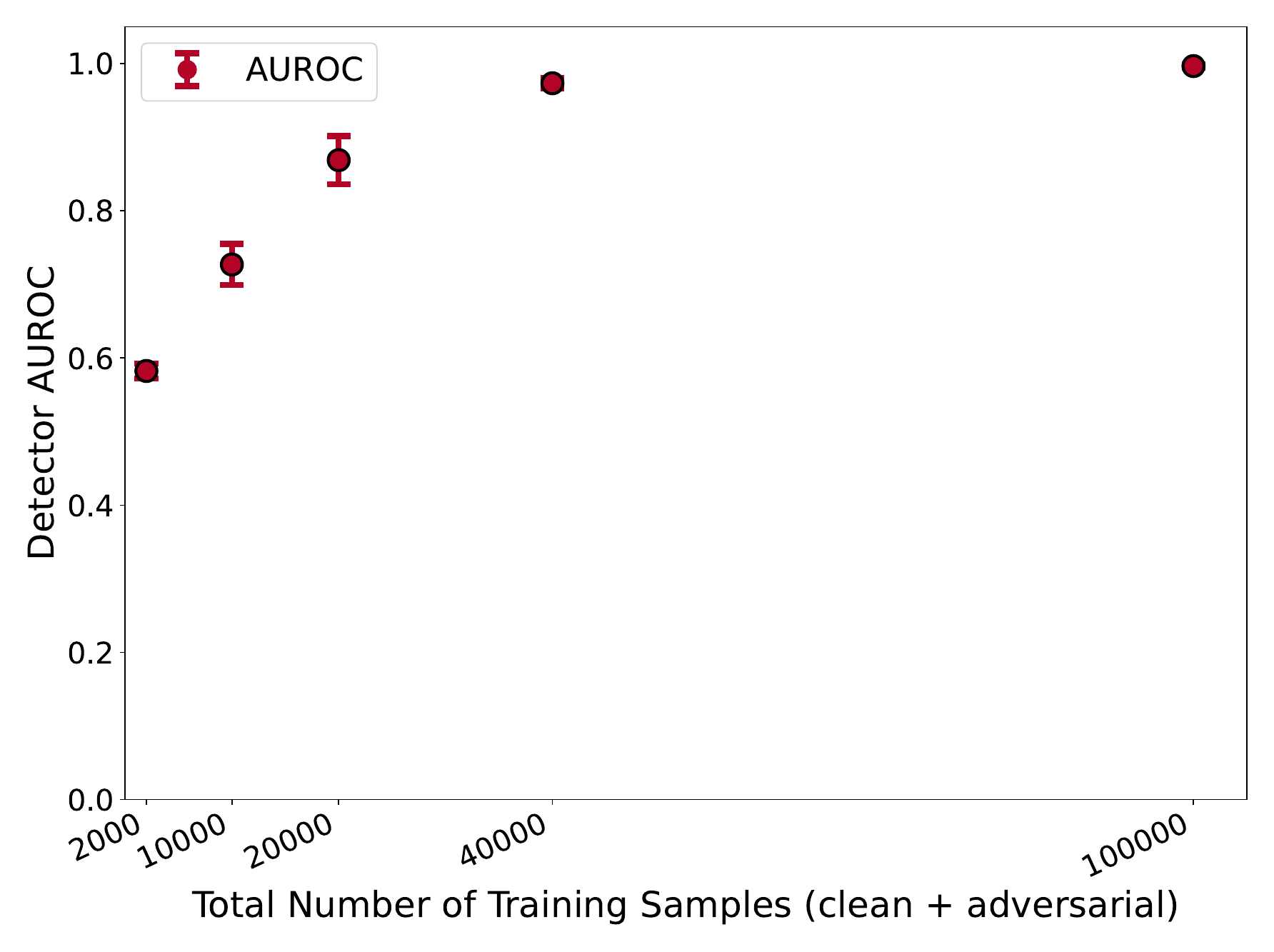}
     \caption{Area under the Receiver-Operating-Curve for varying sample sizes of training events used for the Pearson correlation restricted adversarial detector network.}
     \label{fig:Disco_VD_Auroc}
\end{figure}

To this end, Figure \ref{fig:Disco_VD_Auroc} shows the area under the receiver operating characteristic curve (AUROC) of the adversarial detector---trained on Pearson-correlation-based adversaries---as a function of the total number of training samples. In all cases, the detector training sets are constructed using a 50/50 split between clean and adversarial events.

There, we see that the adversarial detector’s performance improves noticeably for both clean and adversarial samples as the total number of training examples increases. This observation explains why the corrected fooling ratios in the low-sample-size studies are substantially higher than those obtained with larger datasets: the adversarial detector has not yet reached performance saturation when trained on a severely limited number of samples.

\begin{figure}[!htb]
     \centering
     \includegraphics[width=0.49\textwidth]{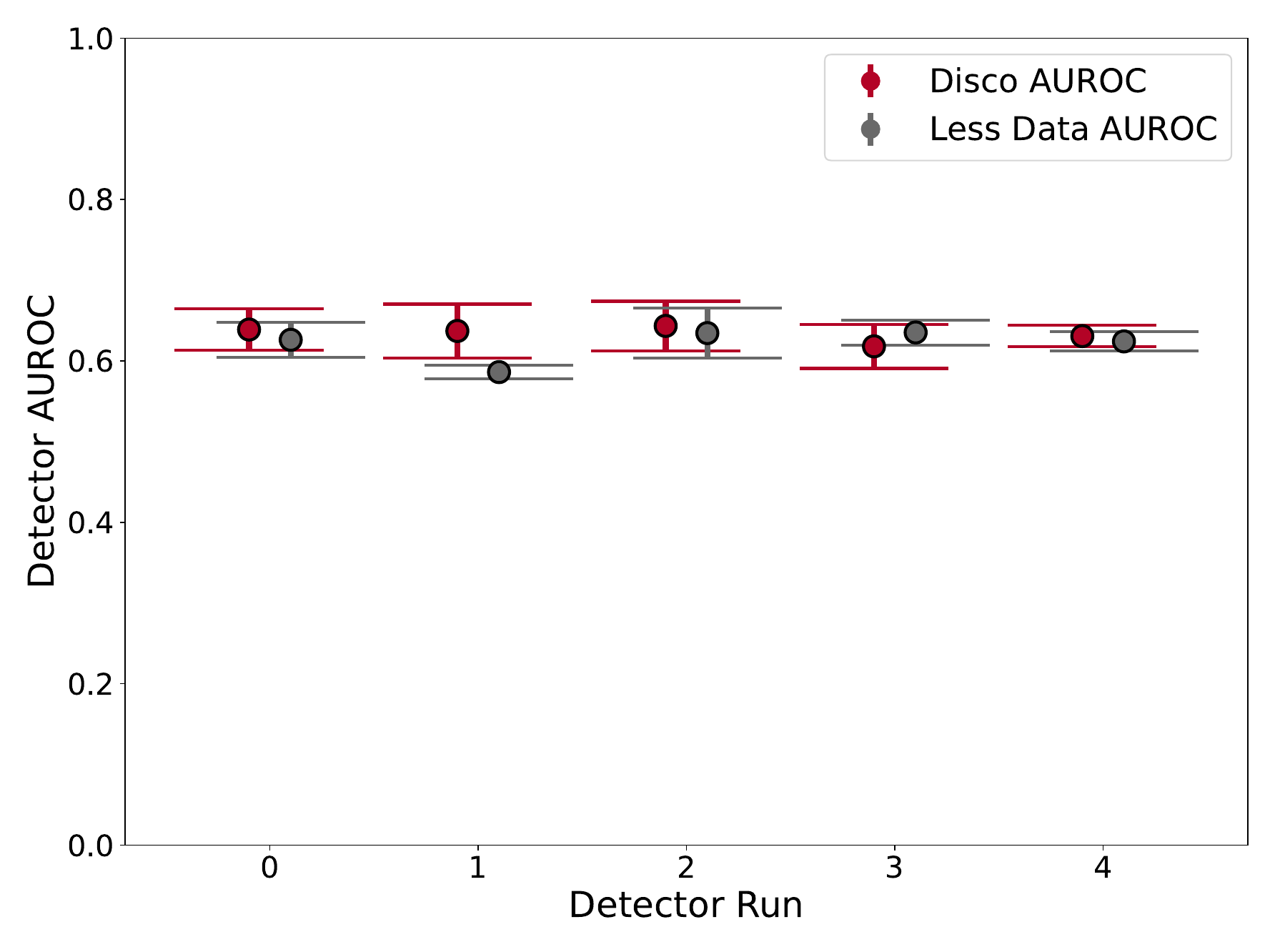}
     \caption{Comparison of the averageg Areas under the Receiver-Operating-Curve for the distance correlation restricted attack (red) and the limited data Pearson restricted attack (grey) across 5 adversarial detector runs.}
     \label{fig:Disco_AUROCComp}
\end{figure}

Comparing adversarial detectors trained on distance-correlation-constrained and Pearson-correlation-constrained adversaries reveals virtually no difference in overall performance, as shown in Figure \ref{fig:Disco_AUROCComp}.

This indicates that the adversarial detector does not struggle to identify distance-correlation-constrained adversaries, despite their more restrictive construction, which could in principle have made them harder to detect. A slightly larger gap between the initial and corrected fooling ratios is indeed observed for the Pearson-based adversaries, suggesting a higher adversarial classification accuracy for the detector trained on Pearson perturbations compared to the distance correlation case.

This improved adversarial detection performance, however, comes at the cost of reduced classification accuracy on clean events relative to the distance-correlation-trained detector. We attribute this behavior to the higher initial fooling ratios achieved by Pearson adversaries. Since the detector is trained exclusively on adversarial samples that successfully fool the baseline model, a higher fooling ratio leads to a larger effective representation of adversarial events in the detector’s training set. This imbalance likely biases the detector slightly toward classifying ambiguous inputs as adversarial.

\section{Toolkit / Workflow}

In light of the potential vulnerabilities demonstrated throughout this work, we propose a practical workflow for incorporating adversarial analyses into studies that rely on simulated HEP data. The central idea is to establish a systematic procedure that not only quantifies the susceptibility of a given model to undetectable adversarial perturbations, but also provides a means of mitigating their impact and estimating the associated uncertainty.

The workflow begins by training a baseline model on clean simulated events, following the standard approach used in a typical HEP analysis. This model can then be used to generate adversarial examples---using the attack proposed in this work---on the training, validation, and test datasets. The clean and adversarial samples are then combined into an augmented dataset and assigned corresponding labels indicating whether they are genuinely clean or adversarially perturbed.

Using this augmented dataset, an Adversarial Detector network---implemented, for instance, as a simple MLP---is trained to distinguish between clean and adversarial events. As shown in our studies, such a detector is capable of learning subtle properties beyond classical low-dimensional statistics, and therefore provides a valuable diagnostic tool for identifying events that behave adversarially with respect to a given predictive model.

After training the detector, one can analyze both the clean and adversarial events in the same fashion as in our experiments. In particular, the fooling ratio of the adversarial events with respect to the baseline model can be compared before and after removing events that the detector identifies as adversarial. This allows a quantification of how much of the model’s susceptibility to adversarial examples can be mitigated---and consequently how the estimated upper-bound on the model’s systematic uncertainty can be reduced---through the use of the detector.

\begin{figure*}[!htb]
     \centering
     \includegraphics[width=0.99\textwidth]{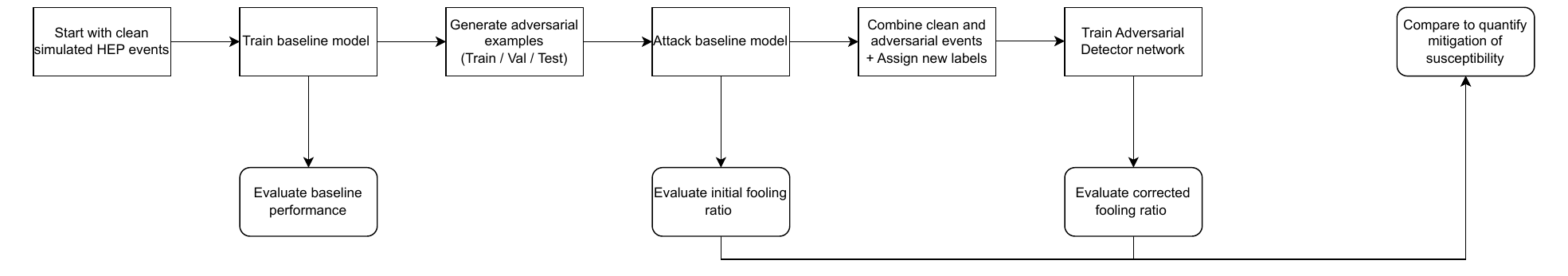}
     \caption{Flowchart showing the proposed workflow.}
     \label{fig:WorkflowFlowchart}
\end{figure*}

A flowchart showing this workflow is shown in Figure \ref{fig:WorkflowFlowchart}.

To support adoption of this workflow, we provide an accompanying \href{https://github.com/TSaala/CONSERVAttack}{GitHub repository} containing a working example of the proposed attack, along with a reference implementation of the full pipeline.

\section{Limitations and Outlook}

In the context of robustness studies, there are numerous promising vectors for further investigation, many of which could improve the estimation of the upper bounds on systematic uncertainties associated with HEP deep learning networks. Furthermore, the adversarial examples constructed in this work open up opportunities to apply---or to develop---methods from explainable AI. Such tools could not only help to better understand why our networks behave as they do, but may also provide insight into potential shortcomings or mismodellings in current HEP simulation pipelines. We believe that a systematic exploration of adversarial directions in feature space could eventually serve as a diagnostic instrument for both ML models and event generators.

Additionally, we believe that the attack proposed could be optimized further. From a computational perspective, more efficient approximations or optimization heuristics may substantially reduce runtime without sacrificing attack performance. Beyond computational efficiency, there is likely room for hyperparameter optimization for each dataset studied here, possibly resulting in even stronger attacks while achieving an even more precise preservation of marginal distributions and their correlations.

A clear limitation of this approach is its reliance on sufficiently large statistics to reliably match correlations and distributions. In low-statistics regimes, the optimization becomes unstable and the attack less well-defined. Moreover, the method as constructed requires continuous or pseudo-continuous features; datasets dominated by categorical variables are not natural candidates for this approach.

\section{Conclusions}

In this work, we introduce the novel adversarial attack CONSERVAttack designed for deep learning models in High Energy Physics. The attack is optimized to produce inputs that induce erroneous model predictions while keeping perturbations to two key statistical properties of the data---marginal distributions and inter-feature correlations---within their respective uncertainty bounds. This design renders the resulting adversarial events virtually invisible to traditional data validation techniques applied in simulation-data comparisons.  

We apply this attack to two common HEP tasks---jet tagging and a signal-background classification problem---demonstrating its ability to achieve high fooling ratios with minimal statistical distortion. Beyond its use as an attack, we show that these events can be repurposed as a data-augmentation tool, improving model performance in low-data scenarios for both tasks.

Conversely, with respect to adversarial defenses, we explore two complementary robustness strategies. First, we perform Adversarial Training using attacks generated on the training set. Second, we introduce an Adversarial Detector: a simple binary classification model capable of distinguishing clean from adversarially perturbed samples. Both defenses, when applied correctly, significantly mitigate susceptibility to our attack and thereby reduce the upper bound on model-induced systematic uncertainty.

To further investigate the robustness of the attack’s statistical preservation, we extend CONSERVAttack beyond linear correlation constraints by replacing the Pearson correlation requirement with Distance Correlation, thereby enforcing the preservation of both linear and non-linear feature dependencies. As expected, the stronger constraint makes the construction of adversarial events substantially more challenging, leading to lower fooling ratios. However, the adversarial detector’s performance remains largely unaffected, indicating that it can still reliably distinguish these more tightly constrained adversaries from clean events. This suggests that the effectiveness of the Pearson-constrained attack cannot be attributed solely to the exploitation of non-linear correlations, but instead reflects a broader vulnerability of the model under statistically consistent perturbations.

Finally, we leverage the Adversarial Detector to conduct further experiments, attempting to understand whether there exist simulated events exhibiting "pseudo-adversarial" behavior, and how the performance of the detector generalizes to clean real data. These studies reveal that while most clean events behave consistently across multiple detector runs, a statistically significant subset mimics adversarial characteristics. Furthermore, the detector's performance transfers remarkably well to real events. 

We propose a workflow to evaluate the vulnerability of machine learning applications in HEP and beyond to unknown effects that may lead to adversarial attacks. We propose that if the corrected fooling ratio, after applying an adversarial detector, lies within the systematic uncertainties bounds derived from physical sources, and if the fraction of adversarial samples detected in data is consistent with that observed in simulation, then no additional uncertainty needs to be assigned to potential adversarial effects. However, if the above conditions are not satisfied, we propose that further investigations should be undertaken to identify potential unaccounted sources of discrepancies between simulation and data. If such studies do not reveal physical origins for the adversarial behavior, the assignment of additional uncertainties should be considered.

\section*{Acknowledgement}
This work has been funded in the context of the AISafety Project, funded by the BMBF under the grant proposal 05D23UM1 and supported in the context of the Clusters of Excellence EXC 3037/533607693 \enquote{Dynaverse} and EXC~3107/533766364 \enquote{Color meets Flavor} under Germany's Excellence Strategy. We thank Markus Prim for very helpful discussions in the context of the example in Section~\ref{sec:donut}. Our work would have been impossible without the services provided by the CERN open data initiative \texttt{openscience.cern} and without the contributions from the LHC experiments. 



\bibliography{./Bibliography}

\begin{appendices}

\section{Attack Optimization}
\label{App:Optimization}

\subsection{Jensen Shannon Distance Calculation}
\label{App:JSDOptimization}

When evaluating the effect of small perturbations on feature histograms (as required for Jensen--Shannon distance calculations), it is not necessary to recompute the entire histogram from scratch for each candidate change. Let $n$ denote the number of events and $K$ the number of histogram bins for a given feature. If a single entry $x_{ij}$ in the dataset is modified, only the bin counts corresponding to the old and new values are affected, while all other bins remain unchanged. Specifically, the histogram can be updated exactly by decrementing the count in the bin containing the old value and incrementing the count in the bin containing the new value. This procedure is mathematically exact because the histogram is simply a count of occurrences in each bin, and a single change only affects the relevant bins.

From a computational perspective, this incremental approach reduces the per-candidate complexity from $O(n)$---required to recompute the histogram over all $n$ samples---to $O(1)$, since only two bins are updated per candidate. This makes it highly efficient for evaluating large numbers of candidate changes, especially in high-dimensional or large-sample settings.

\begin{align}
k_{\text{old}} &:= \text{bin}\!\left(x_{ij}^{\text{old}}\right), \\
k_{\text{new}} &:= \text{bin}\!\left(x_{ij}^{\text{new}}\right), \\
h_{k}^{\text{new}} &:= 
\begin{cases}
h_{k}^{\text{old}} - \frac{1}{n}, & \text{if } k = k_{\text{old}}, \\
h_{k}^{\text{old}} + \frac{1}{n}, & \text{if } k = k_{\text{new}}, \\
h_{k}^{\text{old}}, & \text{otherwise}.
\end{cases}
\end{align}

where $h_k$ denotes the normalized count in bin $k$. This update can be efficiently vectorized for all features and candidate changes, enabling rapid batch evaluation of Jensen--Shannon distances on the GPU.

\subsection{Pearson Correlation Matrix Calculation}
\label{App:PearsonOptimization}

When evaluating the effect of small perturbations on the Pearson correlation matrix, it is not necessary to recompute the entire matrix from scratch for each candidate change. Let $n$ denote the number of events and $d$ the number of features in the dataset. If a single entry in the dataset is modified, the corresponding feature mean can be updated directly, followed by an update of the feature variance, and finally an update of the covariance between the modified feature and all other features. By introducing intermediate variables representing deviations from the old and new means, the covariance update can be expressed compactly and exactly as the difference of products of these deviations. The final correlation entries are then obtained by normalizing the updated covariances by the updated standard deviations. This procedure is mathematically exact because the Pearson correlation depends solely on the means, variances, and covariances of the features, all of which are correctly adjusted. From a computational perspective, this incremental approach reduces the per-candidate complexity from $O(d^2 n)$---required to recompute all pairwise covariances from scratch---to $O(d)$, making it highly efficient for evaluating large numbers of small changes.

\begin{align}
\Delta_j &:= x_{\text{new}} - x_{\text{old}}, \nonumber\\
\delta_j^{\text{old}} &:= x_{\text{old}} - \mu_j^{\text{old}}, \quad
\delta_j^{\text{new}} := x_{\text{new}} - \mu_j^{\text{new}}, \nonumber\\
\delta_k^{\text{old}} &:= x_{ik} - \mu_k^{\text{old}}, \quad
\delta_k^{\text{new}} := x_{ik} - \mu_k^{\text{new}}, \nonumber\\
\mu_j^{\text{new}} &:= \mu_j^{\text{old}} + \frac{\Delta_j}{n}, \nonumber\\
\sigma_j^{2, \text{new}} &:= \sigma_j^{2, \text{old}} + 
   \frac{ (\delta_j^{\text{new}})^2 - (\delta_j^{\text{old}})^2 }{n-1}, \nonumber\\
\text{cov}_{jk}^{\text{new}} &:= \text{cov}_{jk}^{\text{old}} + 
   \frac{ \delta_j^{\text{new}} \, \delta_k^{\text{new}} 
          - \delta_j^{\text{old}} \, \delta_k^{\text{old}} }{n-1}, \nonumber\\
\rho_{jk}^{\text{new}} &:= \frac{\text{cov}_{jk}^{\text{new}}}{\sigma_j^{\text{new}} \, \sigma_k^{\text{new}}}.
\end{align}

\section{Control Region Attack}
\label{App:CRAttack}

In this section, we consider a physically motivated variant of the CONSERVAttack. To define the control and signal regions, we identify the most discriminating feature by performing an exhaustive search over all input variables for the optimal single-variable decision boundary, i.e., the feature and threshold that maximize classification accuracy using a single cut. This procedure selects the reconstructed invariant mass from the Missing Mass Calculator (\texttt{DER\_mass\_MMC}) as the best separator. The control region is defined as the subset of events below this optimal threshold, corresponding to a background-enriched phase-space region where the distributions and correlations are well understood. The signal region comprises all remaining events above the threshold.

In this variant, the marginal distributions and linear correlations are constrained only in the control region. In the signal region---where the corresponding distributions and correlations are typically unknown---large, unconstrained perturbations are allowed.

The effectiveness of the attack is evaluated as follows. For the Fooling Ratio, we consider three subsets: the control region, the signal region, and the full dataset. For $D_{\mathrm{JS}}$ and $\Delta_F$, we restrict the comparison to the control region before and after the attack, as this is the only region in which statistical consistency can be reliably validated.

\begin{figure}[!htb]
\centering
\includegraphics[width=0.49\textwidth]{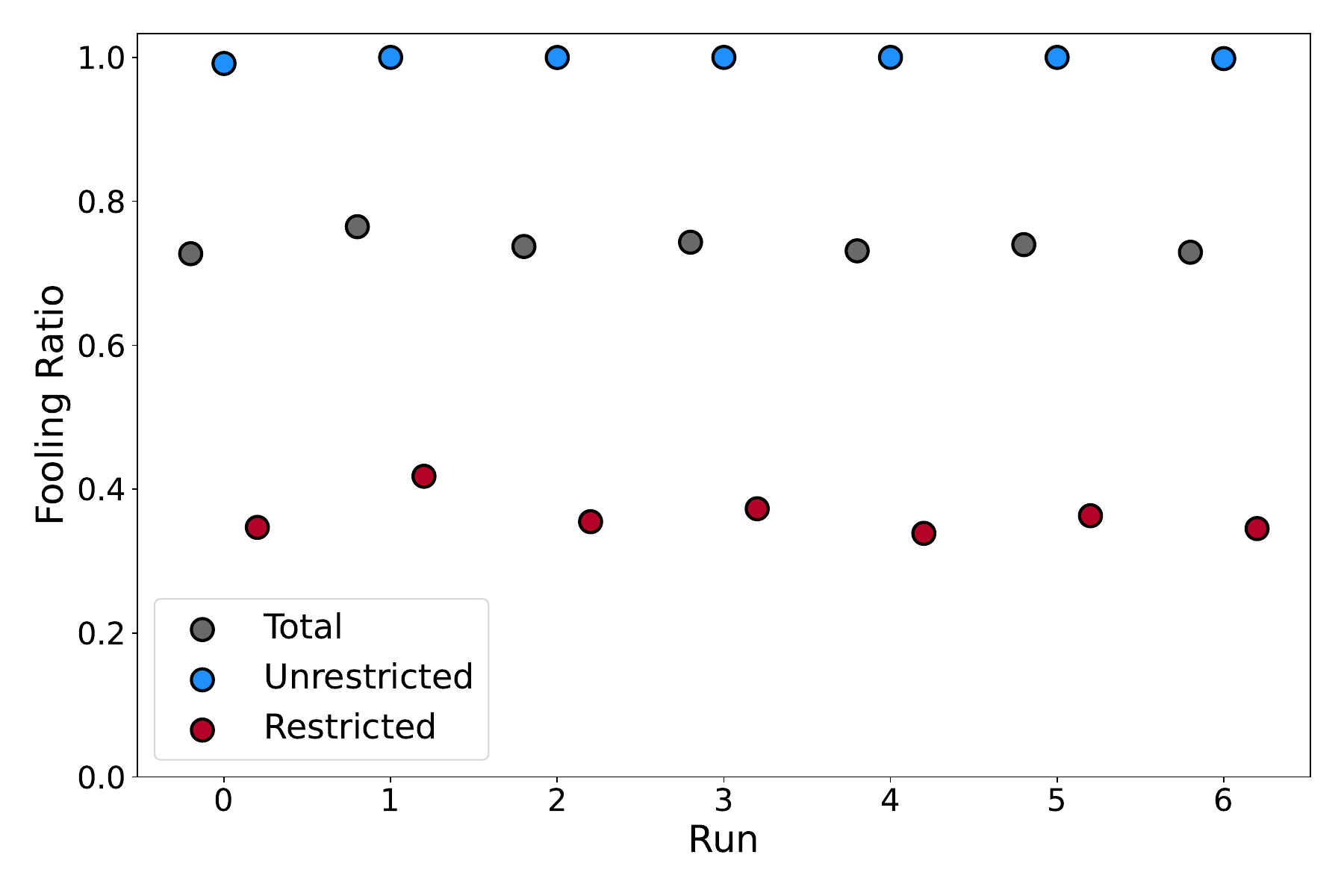}
\caption{Fooling ratios of the restricted attack on the Higgs dataset over 10 independent runs. The grey points indicate the total fooling ratio across all events, the blue points correspond to the signal (unrestricted) region, and the red points to the control (restricted) region.}
\label{fig:RestrictedFR}
\end{figure}

\begin{figure}[!htb]
\centering
\includegraphics[width=0.49\textwidth]{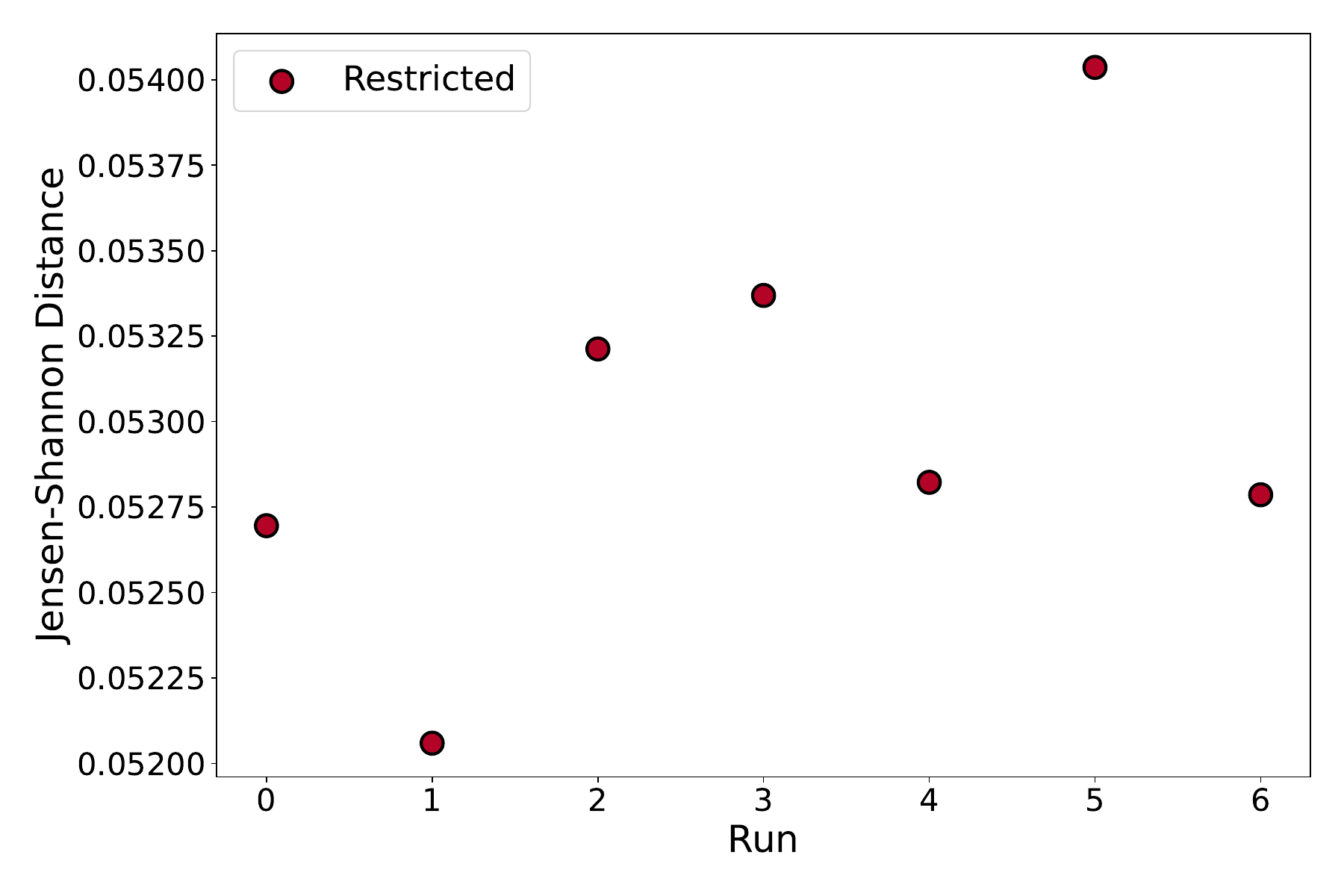}
\caption{Average Jensen--Shannon distance of the adversarial attack evaluated on the control (restricted) region of the Higgs dataset over 10 runs.}
\label{fig:RestrictedJSD}
\end{figure}

\begin{figure}[!htb]
\centering
\includegraphics[width=0.49\textwidth]{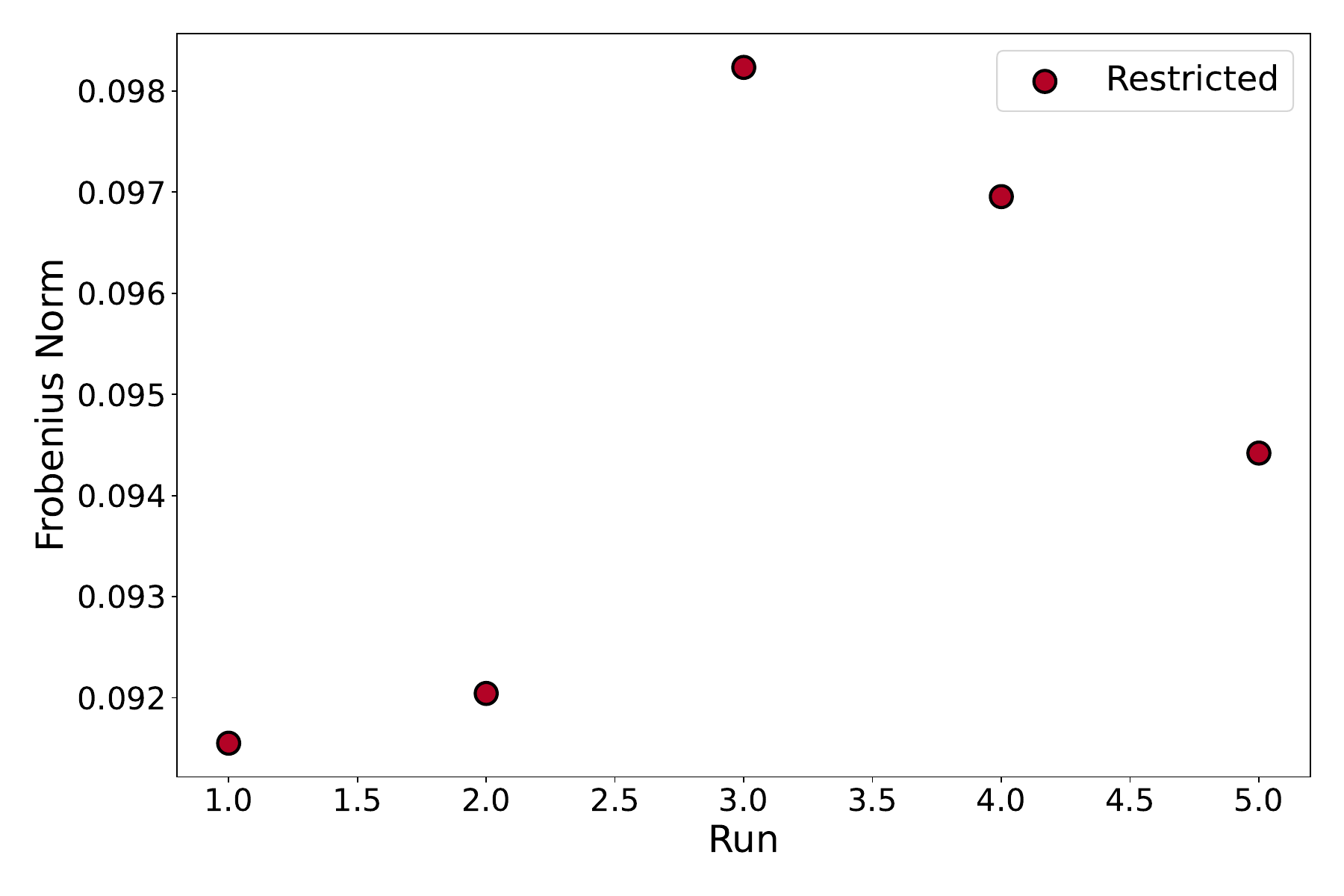}
\caption{Average Frobenius norm of the adversarial attack evaluated on the control (restricted) region of the Higgs dataset over 10 runs.}
\label{fig:RestrictedFN}
\end{figure}

The results are shown in Figures \ref{fig:RestrictedFR}, \ref{fig:RestrictedJSD}, and \ref{fig:RestrictedFN}.

For the Fooling Ratio, we observe values close to 1 in the signal (unrestricted) region, as expected given that large perturbations are permitted. In contrast, the Fooling Ratio in the control (restricted) region is noticeably lower, reflecting the constraints imposed on the marginal distributions and linear correlations. Nevertheless, when evaluated over the full dataset, the combined attack achieves a Fooling Ratio of nearly 0.8 across all runs.

For $D_{\mathrm{JS}}$ and $\Delta_F$---evaluated exclusively in the control region---we obtain small values (approximately 0.05 for $D_{\mathrm{JS}}$ and 0.09 for $\Delta_F$). Consistent with the previous attack setting, this indicates that the perturbations introduced in the constrained region remain effectively undetectable with respect to low-dimensional statistical properties. In particular, this suggests that such an attack would likely pass standard validation checks performed in the control region.

\section{Datasets and Networks}
\label{App:DataNet}

\subsection{Higgs}

As input for the Multi Layer Perceptron used to tackle the Higgs signal--background
classification task popularized by the corresponding Kaggle challenge, we use the
following 30 features:
\texttt{DER\_mass\_MMC},
\texttt{DER\_mass\_transverse\_met\_lep},
\texttt{DER\_mass\_vis},
\texttt{DER\_pt\_h},
\texttt{DER\_deltaeta\_jet\_jet},
\texttt{DER\_mass\_jet\_jet},
\texttt{DER\_prodeta\_jet\_jet},
\texttt{DER\_deltar\_tau\_lep},
\texttt{DER\_pt\_tot},
\texttt{DER\_sum\_pt},
\texttt{DER\_pt\_ratio\_lep\_tau},
\texttt{DER\_met\_phi\_centrality},
\texttt{DER\_lep\_eta\_centrality},
\texttt{PRI\_tau\_pt},
\texttt{PRI\_tau\_eta},
\texttt{PRI\_tau\_phi},
\texttt{PRI\_lep\_pt},
\texttt{PRI\_lep\_eta},
\texttt{PRI\_lep\_phi},
\texttt{PRI\_met},
\texttt{PRI\_met\_phi},
\texttt{PRI\_met\_sumet},
\texttt{PRI\_jet\_num},
\texttt{PRI\_jet\_leading\_pt},
\texttt{PRI\_jet\_leading\_eta},
\texttt{PRI\_jet\_leading\_phi},
\texttt{PRI\_jet\_subleading\_pt},
\texttt{PRI\_jet\_subleading\_eta},
\texttt{PRI\_jet\_subleading\_phi},
and \texttt{PRI\_jet\_all\_pt}.
Features prefixed with \texttt{PRI} are primitive raw quantities describing the bunch
collision as measured directly by the detector, while features prefixed with
\texttt{DER} are quantities derived from the primitive features.

The hyperparameter setup for the baseline classifier is shown in Table \ref{table:HiggsBaselineArchitecture} and the adversarial detector setup in Table \ref{table:HiggsDetectorArchitecture}.

\begin{table}[h!]
\caption{Architecture of the baseline signal-background classifier for the Higgs dataset.}
\begin{tabular}{ |p{2.5cm}|p{1.1cm}|p{2.9cm}| }
 \hline
 \multicolumn{3}{|c|}{\textbf{Higgs - Signal-Background Classifier}} \\
 \hline
 \textbf{Layer} & \textbf{Nodes} & \textbf{Activation Function} \\
 \hline \hline
 Input Layer & 30 & - \\
 \hline
 Dense & 300 & ReLU \\
 \hline
 BatchNormalization & - & - \\
 \hline
 Dense & 102 & ReLU \\
 \hline
 BatchNormalization & - & - \\
 \hline
 Dense & 12 & ReLU \\
 \hline
 BatchNormalization & - & - \\
 \hline
 Dense & 6 & ReLU \\
 \hline
 BatchNormalization & - & - \\
 \hline
 Dense (Output) & 1 & Sigmoid \\
 \hline
\end{tabular}
\label{table:HiggsBaselineArchitecture}
\end{table}

\begin{table}[h!]
\caption{Architecture of the adversarial detector MLP for the Higgs dataset.}
\begin{tabular}{ |p{2.5cm}|p{1.1cm}|p{2.9cm}| }
 \hline
 \multicolumn{3}{|c|}{\textbf{Higgs - Adversarial Detector Model}} \\
 \hline
 \textbf{Layer} & \textbf{Nodes} & \textbf{Activation Function} \\
 \hline \hline
 Input Layer & 30 & - \\
 \hline
 Dense & 300 & ReLU \\
 \hline
 BatchNormalization & - & - \\
 \hline
 Dense & 102 & ReLU \\
 \hline
 BatchNormalization & - & - \\
 \hline
 Dense & 12 & ReLU \\
 \hline
 BatchNormalization & - & - \\
 \hline
 Dense & 6 & ReLU \\
 \hline
 BatchNormalization & - & - \\
 \hline
 Dense (Output) & 1 & Sigmoid \\
 \hline
\end{tabular}
\label{table:HiggsDetectorArchitecture}
\end{table}

\subsection{TT vs. WW Jets}

As input for this TopoDNN inspired model on the TT vs. WW Jet classification task, we take $p_T$, $\eta$, and $\phi$ of the first 30 jet constituents, sorted by their momentum. However, we leave out $\eta$ and $\phi$ of the first constituent, as well as $\eta$ of the second constituent, as these take always the same value due to the pre-processing applied, resulting in a total of 87 input features. A more detailed overview of the variables can be found in the original paper \cite{10.21468/SciPostPhys.7.1.014}.

The hyperparameter setup for the baseline classifier is shown in Table \ref{table:TopoDNNBaselineArchitecture} and the adversarial detector setup in Table \ref{table:TopoDNNDetectorArchitecture}.

\begin{table}[h!]
\caption{Architecture of the baseline signal-background classifier for the TT vs. WW Jets dataset.}
\begin{tabular}{ |p{2.5cm}|p{1.1cm}|p{2.9cm}| }
 \hline
 \multicolumn{3}{|c|}{\textbf{TT vs. WW - Jet Tagger}} \\
 \hline
 \textbf{Layer} & \textbf{Nodes} & \textbf{Activation Function} \\
 \hline \hline
 Input Layer & 87 & - \\
 \hline
 Dense & 300 & ReLU \\
 \hline
 BatchNormalization & - & - \\
 \hline
 Dense & 102 & ReLU \\
 \hline
 BatchNormalization & - & - \\
 \hline
 Dense & 12 & ReLU \\
 \hline
 BatchNormalization & - & - \\
 \hline
 Dense & 6 & ReLU \\
 \hline
 BatchNormalization & - & - \\
 \hline
 Dense (Output) & 1 & Sigmoid \\
 \hline
\end{tabular}
\label{table:TopoDNNBaselineArchitecture}
\end{table}

\begin{table}[h!]
\caption{Architecture of the adversarial detector MLP for the TT vs. WW Jets dataset.}
\begin{tabular}{ |p{2.5cm}|p{1.1cm}|p{2.9cm}| }
 \hline
 \multicolumn{3}{|c|}{\textbf{TT vs. WW - Adversarial Detector Model}} \\
 \hline
 \textbf{Layer} & \textbf{Nodes} & \textbf{Activation Function} \\
 \hline \hline
 Input Layer & 87 & - \\
 \hline
 Dense & 300 & ReLU \\
 \hline
 BatchNormalization & - & - \\
 \hline
 Dense & 102 & ReLU \\
 \hline
 BatchNormalization & - & - \\
 \hline
 Dense & 12 & ReLU \\
 \hline
 BatchNormalization & - & - \\
 \hline
 Dense & 6 & ReLU \\
 \hline
 BatchNormalization & - & - \\
 \hline
 Dense (Output) & 1 & Sigmoid \\
 \hline
\end{tabular}
\label{table:TopoDNNDetectorArchitecture}
\end{table}

\clearpage

\section{Additional Result Plots}
\subsection{Attack Results (TT vs. WW Jets)}
\label{App:AttackRes}

\begin{figure}[!htb]
     \centering
     \includegraphics[width=0.49\textwidth]{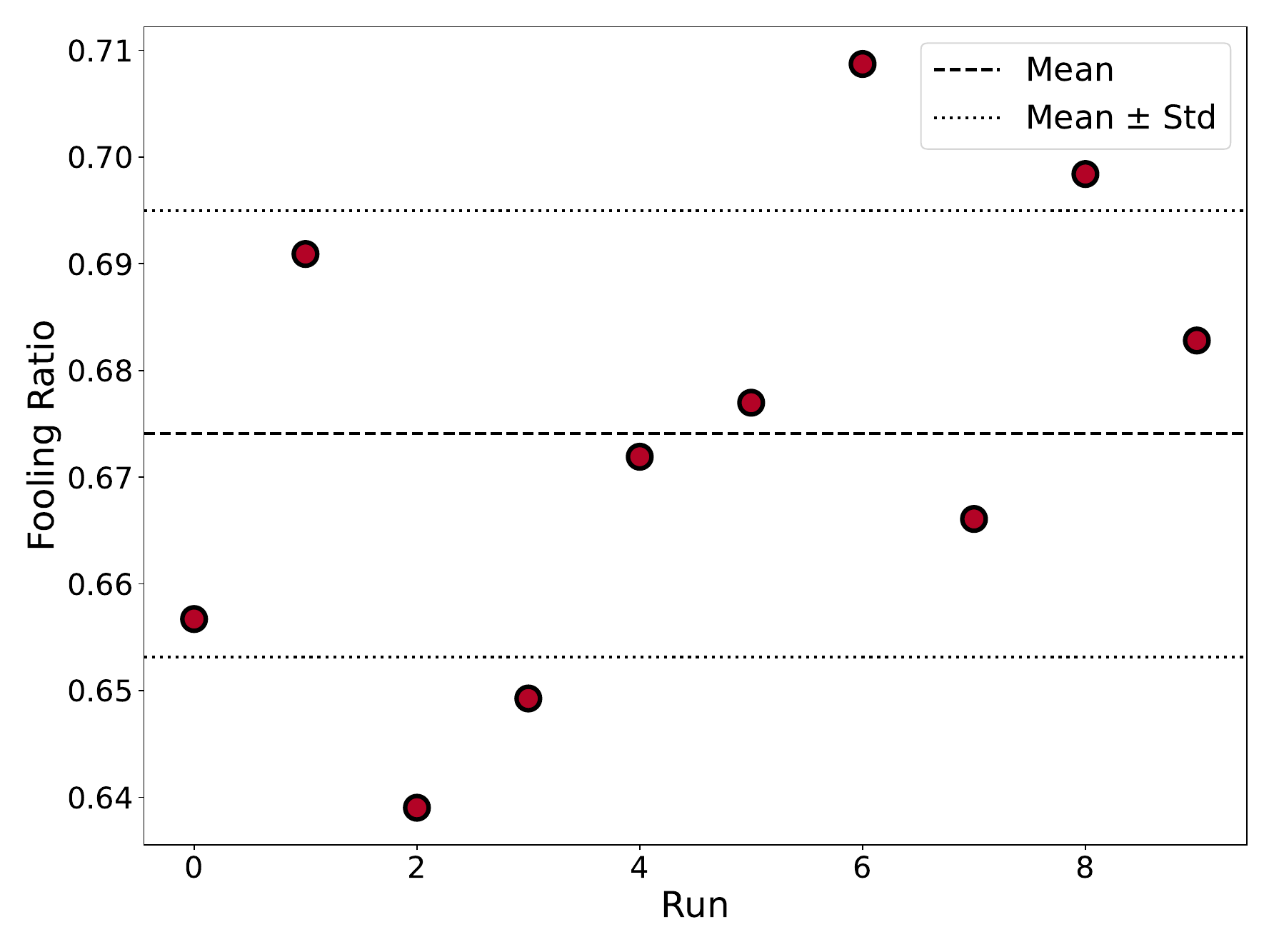}
     \caption{Average fooling ratio of the adversarial attack on the TT vs. WW dataset over 10 runs. The average performance of the attack is shown as a dashed line, and its standard deviation as a dotted line.}
     \label{fig:TTvsWW_AttackFR}
\end{figure}

\begin{figure}[!htb]
     \centering
     \includegraphics[width=0.49\textwidth]{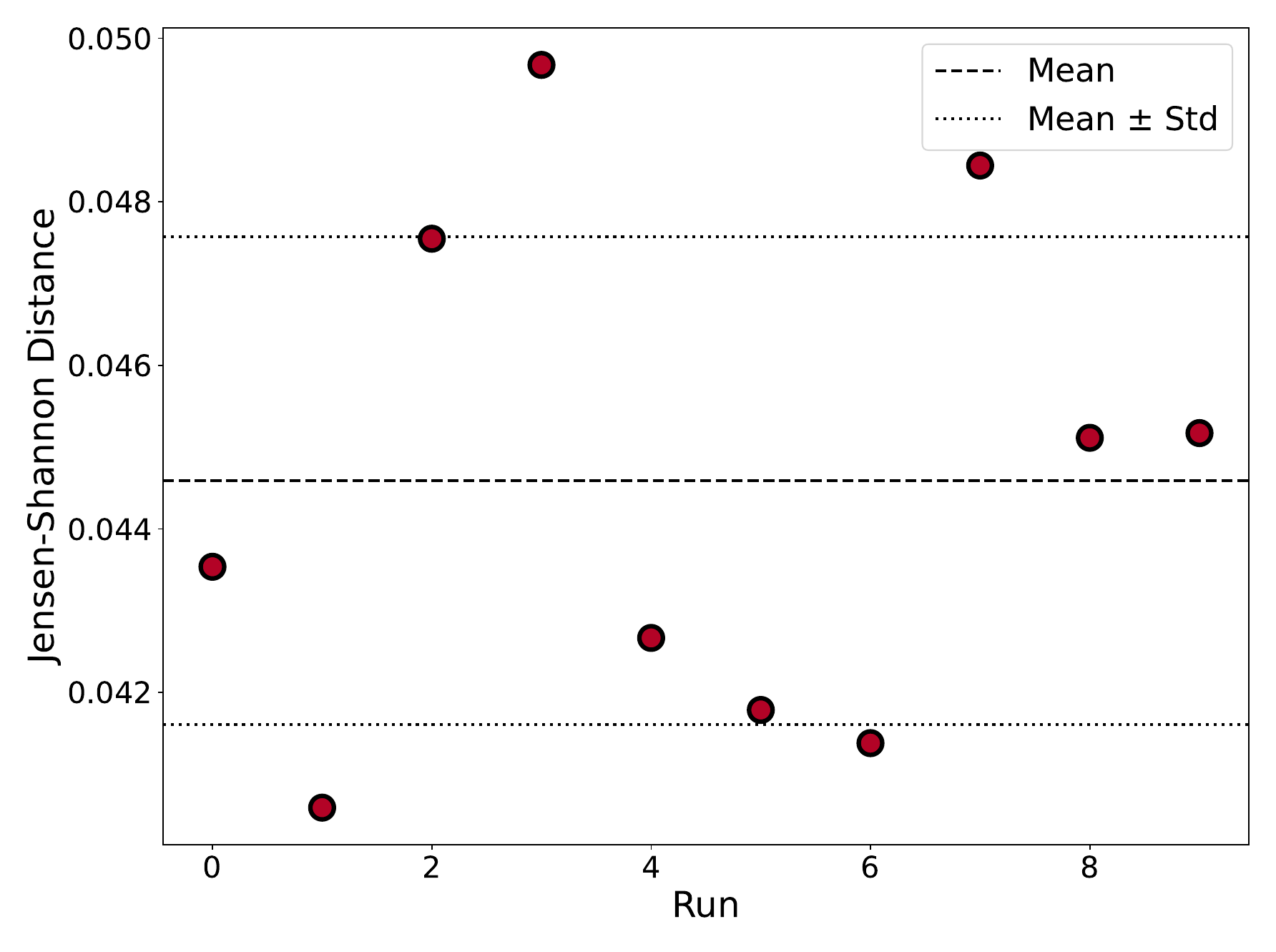}
     \caption{Average Jensen Shannon distance of the adversarial attack on the TT vs. WW dataset over 10 runs. The average performance of the attack is shown as a dashed line, and its standard deviation as a dotted line.}
     \label{fig:TTvsWW_AttackJSD}
\end{figure}

\begin{figure}[!htb]
     \centering
     \includegraphics[width=0.49\textwidth]{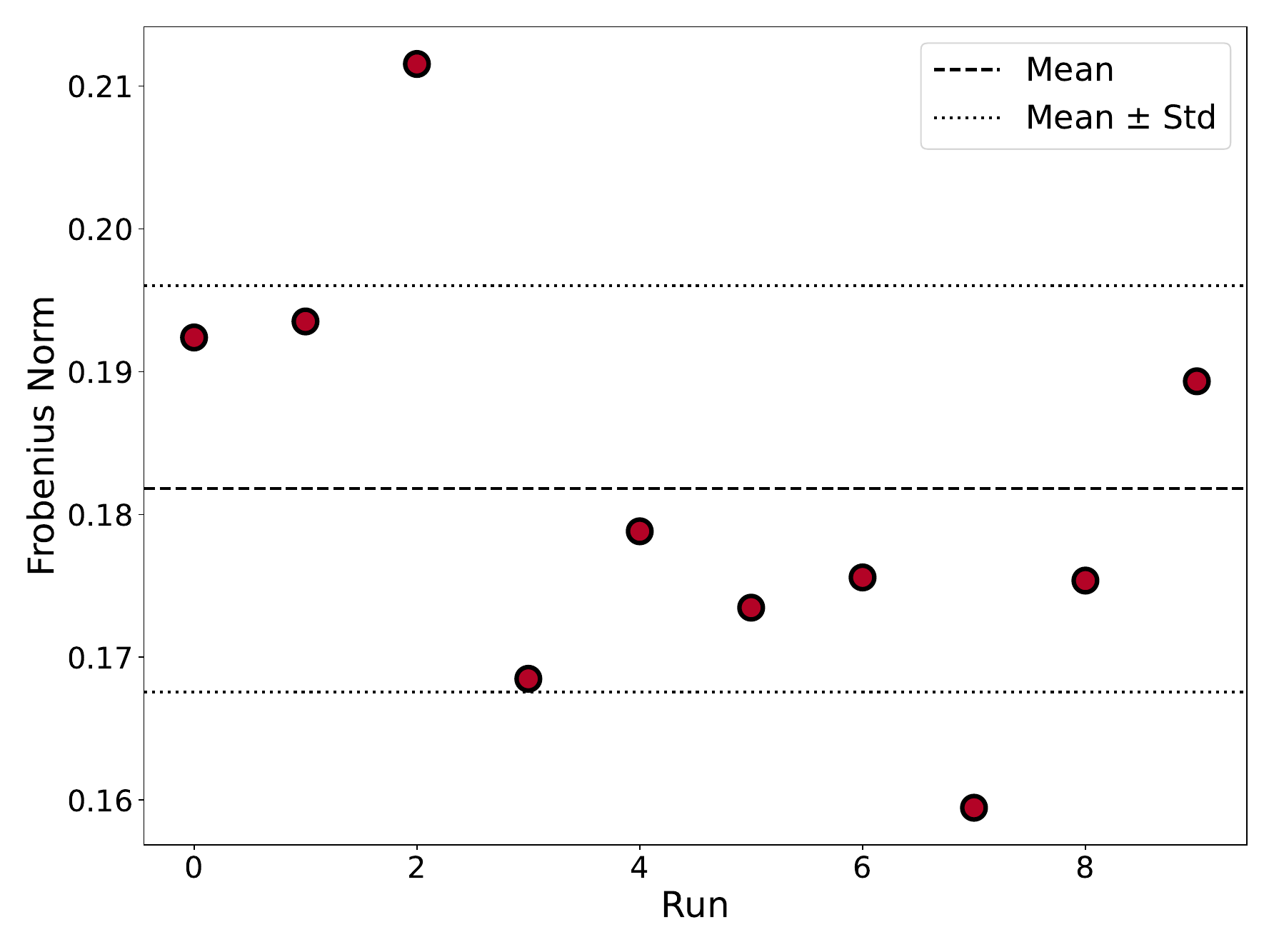}
     \caption{Average Frobenius Norm of the adversarial attack on the TT vs. WW dataset over 10 runs. The average performance of the attack is shown as a dashed line, and its standard deviation as a dotted line.}
     \label{fig:TTvsWW_AttackFN}
\end{figure}

\subsection{Augmentation Results (TT vs. WW Jets)}
\label{App:AugRes}

\begin{figure}[!htb]
     \centering
     \includegraphics[width=0.49\textwidth]{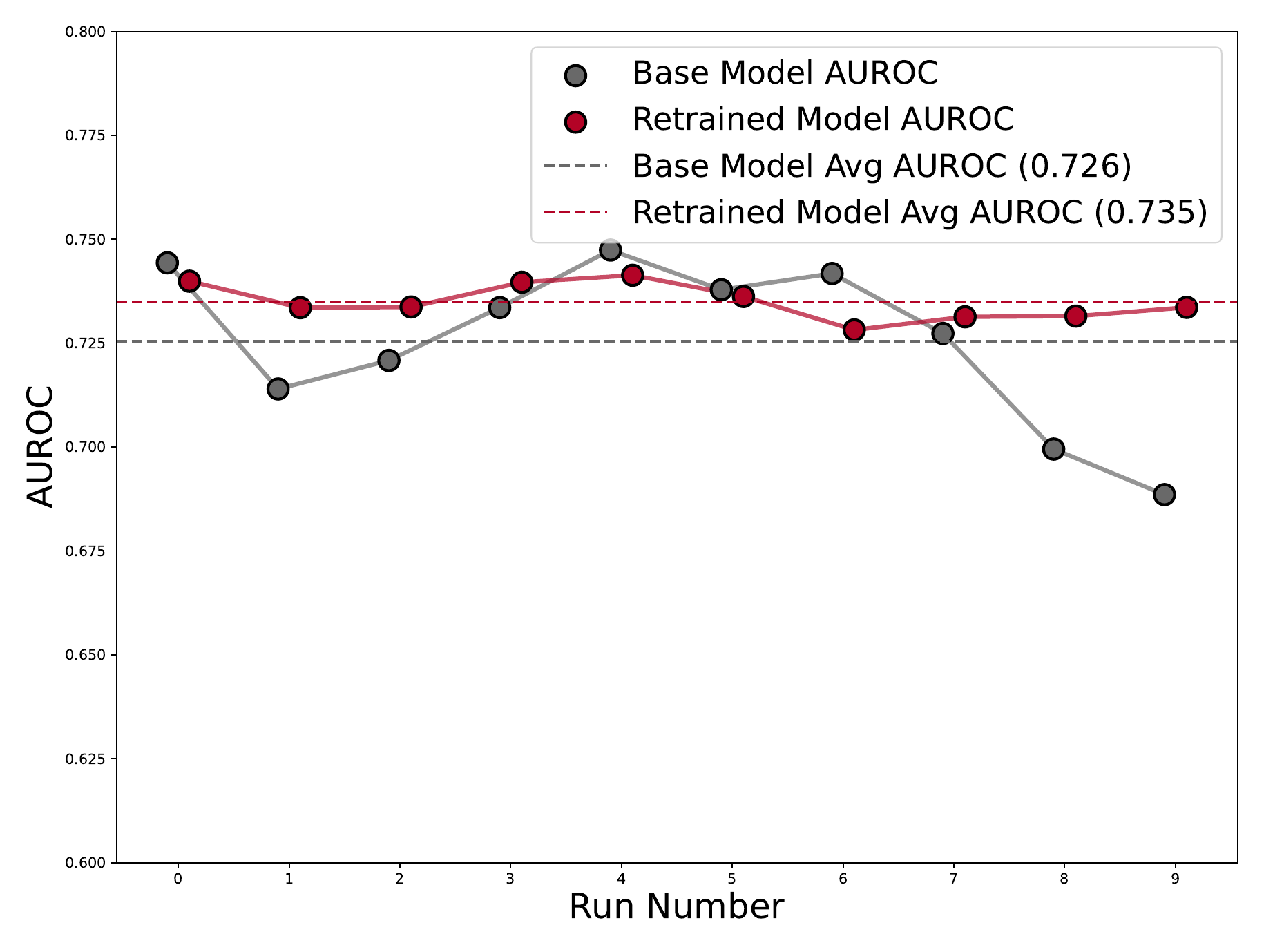}
     \caption{Comparison of the areas under the Receiver-Operating-Curves---on the base model (grey) and the adversarially augmented model (red)---on the Higgs data set over 10 independent runs. The respective average results are shown in dashed lines.}
     \label{fig:TTvsWW_Augmentation}
\end{figure}

\subsection{Robustness Results (TT vs. WW Jets)}
\label{App:RobustRes}

\subsubsection{Adversarial Training}
\label{App:RobustResTrain}

\begin{figure}[!htb]
     \centering
     \includegraphics[width=0.49\textwidth]{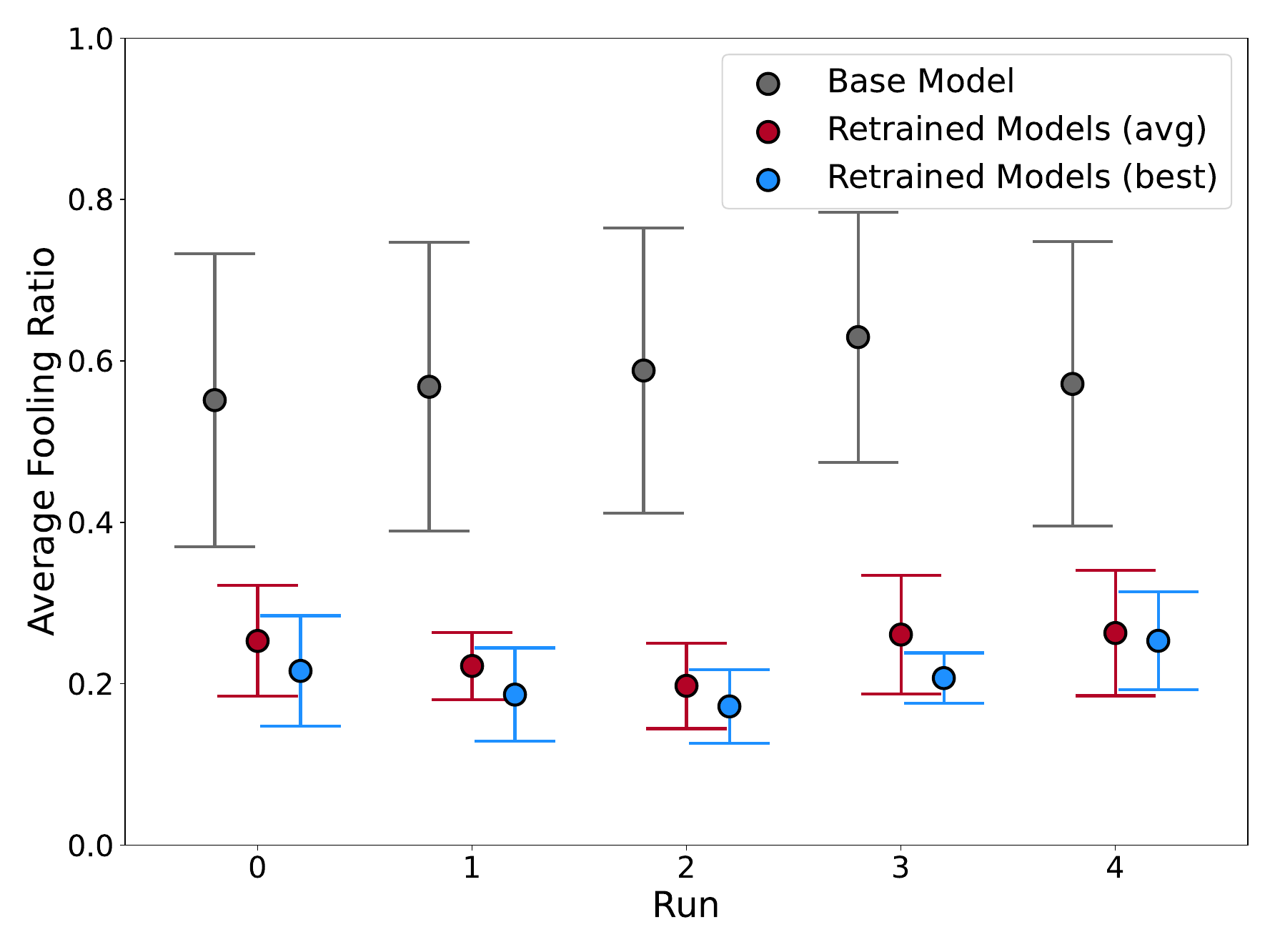}
     \caption{Comparison of the fooling ratios of adversarial attacks on the baseline model (grey) vs the average-case adversarially trained network (red) and the best-case adversarially trained network (blue) on the TT vs. WW dataset across 5 runs.}
     \label{fig:TTvsWW_Retraining}
\end{figure}

\subsubsection{Adversarial Detector}
\label{App:RobustResDetector}

\begin{figure}[!htb]
     \centering
     \includegraphics[width=0.49\textwidth]{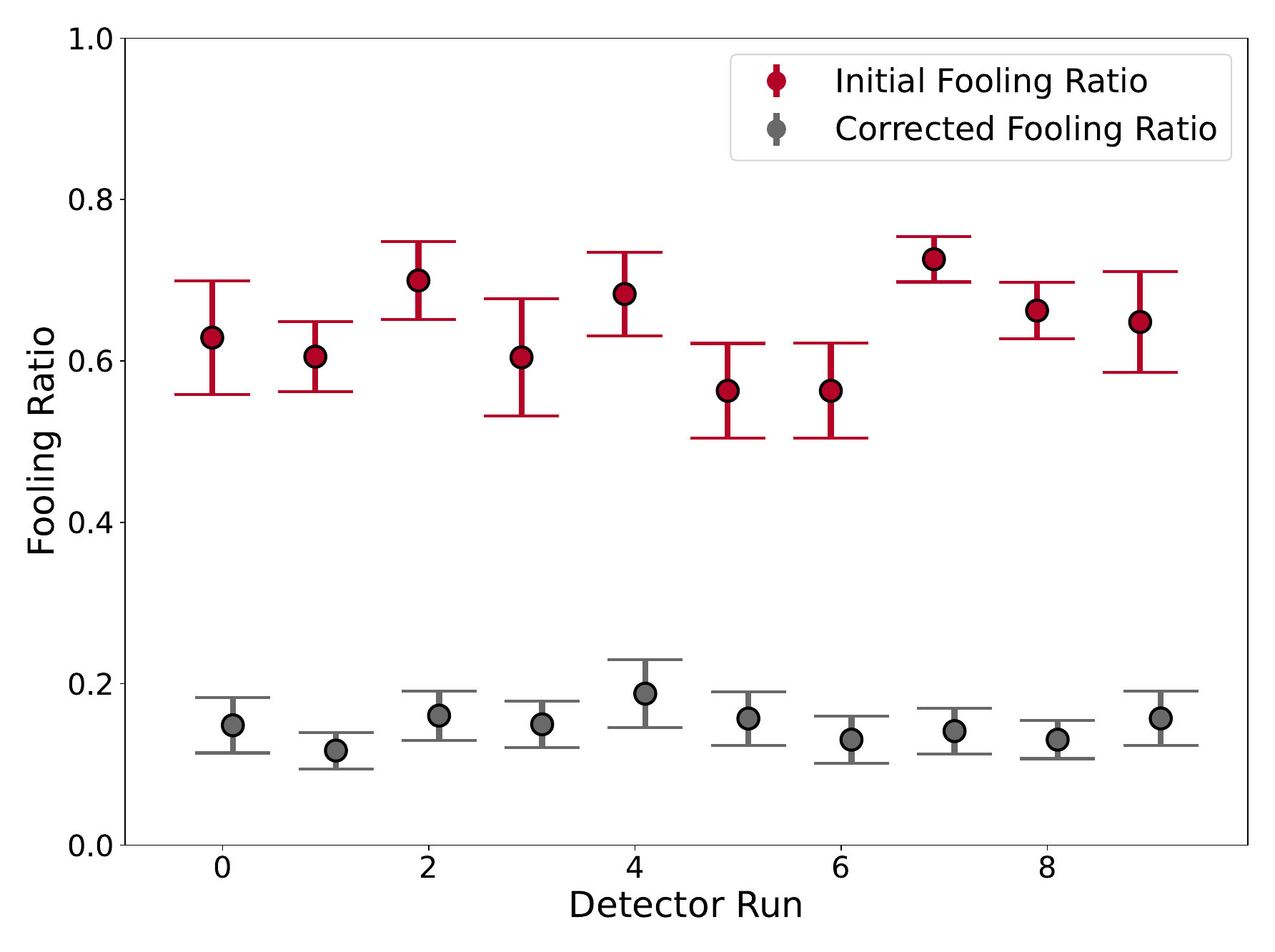}
     \caption{Comparison of the initial fooling ratios (red) and the corrected fooling ratios (grey) found in 10 independent runs of the adversarial detector pipeline on the TT vs. WW dataset.}
     \label{fig:TTvsWW_Detector}
\end{figure}

\clearpage

\section{Attack Parameters}
\label{App:AttackParams}

\begin{table}[h!]
\caption{User-configurable hyperparameters for the adversarial attack.}
\begin{tabular}{ |p{2.2cm}|p{1cm}|p{3cm}| }
 \hline
 \multicolumn{3}{|c|}{\textbf{Adversarial Attack Hyperparameters}} \\
 \hline
 \textbf{Parameter Name} & \textbf{Type} & \textbf{Description} \\
 \hline \hline
x & Numpy Array & Input data to be attacked.  \\
\hline
y & Numpy Array & True labels for the input data. \\
\hline
model & str & Path to the trained model. \\
\hline
min\_change & float & Minimum allowed change per feature in each step. \\
\hline
step & float & Step size for candidate feature changes. \\
\hline
num\_candidates & int/None & Number of random candidate values per feature (overwrites step). \\
\hline
n\_iterations & int & Number of attack iterations to perform. \\
\hline
n\_gpus & int & Number of GPUs to use for parallelization. \\
\hline
num\_bins & int & Number of bins for feature histograms ($D_{\mathrm{JS}}$ calculation). \\
\hline
mask & bool array & Boolean array indicating which samples to target for adversarial modification. \\
\hline
alpha & float & Weight for Jensen-Shannon distance in the cost function. \\
\hline
beta & float & Weight for correlation change in the cost function. \\
\hline
\makecell[l]{max\_jsd\_\\single\_change} & float & Maximum allowed $D_{\mathrm{JS}}$ for a single feature change. \\
\hline
\makecell[l]{max\_frob\_\\single\_change} & float & Maximum allowed Frobenius norm change for a single feature change. \\
\hline
use\_no\_change & bool & Whether to allow "no change" as a candidate for each feature. \\
\hline
\makecell[l]{optimize\_already\_\\fooled} & bool & Whether to further optimize already fooled samples. \\
\hline
use\_disco & bool & Use DisCo correlation instead of Pearson. \\
\hline
\end{tabular}
\label{table:AttackHyperparameters}
\end{table}

\subsection{Higgs}

\begin{table}[h!]
\caption{Parameter values used for adversarial attack for the Higgs dataset.}
\begin{tabular}{ |p{3.25cm}|p{3.25cm}| }
\hline
 \multicolumn{2}{|c|}{\textbf{Higgs Attack Hyperparameter Choice}} \\
 \hline
 \textbf{Parameter Name} & \textbf{Value} \\
 \hline \hline
min\_change & 0.0005 \\
\hline
step & 0.0005 \\
\hline
n\_iterations & 10 \\
\hline
num\_bins & 200 \\
\hline
alpha & 4.0 \\
\hline
beta & 1.0 \\
\hline
use\_no\_change & True \\
\hline
max\_jsd\_single\_change & 0.006 \\
\hline
max\_frob\_single\_change & 0.0002 \\
\hline
optimize\_already\_fooled & False \\
\hline
use\_disco & False \\
\hline
\end{tabular}
\label{table:HiggsAttackParameterValues}
\end{table}

\begin{table}[h!]
\caption{Parameter values used for the augmentation pipeline for the Higgs dataset.}
\begin{tabular}{ |p{3.25cm}|p{3.25cm}| }
\hline
 \multicolumn{2}{|c|}{\textbf{Higgs Attack Hyperparameter Choice}} \\
 \hline
 \textbf{Parameter Name} & \textbf{Value} \\
 \hline \hline
min\_change & 0.002 \\
\hline
step & 0.006 \\
\hline
n\_iterations & 10 \\
\hline
num\_bins & 200 \\
\hline
alpha & 6.5 \\
\hline
beta & 1.0 \\
\hline
use\_no\_change & True \\
\hline
max\_jsd\_single\_change & 0.01 \\
\hline
max\_frob\_single\_change & 0.001 \\
\hline
optimize\_already\_fooled & False \\
\hline
use\_disco & False \\
\hline
\end{tabular}
\label{table:HiggsAugmentationParameterValues}
\end{table}

\begin{table}[h!]
\caption{Parameter values used for the adversarial detector pipeline for the Higgs dataset for training and test samples.}
\begin{tabular}{ |p{3.25cm}|p{3.25cm}| }
\hline
 \multicolumn{2}{|c|}{\textbf{Higgs Attack Hyperparameter Choice}} \\
 \hline
 \textbf{Parameter Name} & \textbf{Value} \\
 \hline \hline
min\_change & 0.003 \\
\hline
step & 0.02 \\
\hline
n\_iterations & 10 \\
\hline
num\_bins & 100 \\
\hline
alpha & 6.0 \\
\hline
beta & 1.0 \\
\hline
use\_no\_change & True \\
\hline
max\_jsd\_single\_change & 0.01 \\
\hline
max\_frob\_single\_change & 0.0003 \\
\hline
optimize\_already\_fooled & False \\
\hline
use\_disco & False \\
\hline
\end{tabular}
\label{table:HiggsDetectorParameterValues}
\end{table}

\begin{table}[h!]
\caption{Parameter ranges used for generating training adversaries in the iterative retraining pipeline (randomized per retraining iteration).}
\begin{tabular}{ |p{3.25cm}|p{3.25cm}| }
\hline
 \multicolumn{2}{|c|}{\textbf{Higgs Attack Hyperparameter Ranges}} \\
 \hline
 \textbf{Parameter Name} & \textbf{Value / Range} \\
 \hline \hline
min\_change & [0.001, 0.005] (uniform) \\
\hline
step & [0.005, 0.03] (uniform) \\
\hline
n\_iterations & 10 \\
\hline
num\_bins & 100 \\
\hline
alpha & [3.0, 10.0] (uniform) \\
\hline
beta & [0.5, 2.5] (uniform) \\
\hline
use\_no\_change & True \\
\hline
max\_jsd\_single\_change & [0.005, 0.03] (uniform) \\
\hline
max\_frob\_single\_change & [0.0001, 0.001] (uniform) \\
\hline
optimize\_already\_fooled & False \\
\hline
use\_disco & False \\
\hline
\end{tabular}
\label{table:HiggsRetrainingTrainParameterRanges}
\end{table}

\begin{table}[h!]
\caption{Parameter values used for generating test adversaries in the iterative retraining pipeline.}
\begin{tabular}{ |p{3.25cm}|p{3.25cm}| }
\hline
 \multicolumn{2}{|c|}{\textbf{Higgs Attack Hyperparameter Choice (Test)}} \\
 \hline
 \textbf{Parameter Name} & \textbf{Value} \\
 \hline \hline
min\_change & 0.003 \\
\hline
step & 0.01 \\
\hline
n\_iterations & 10 \\
\hline
num\_bins & 100 \\
\hline
alpha & 8.0 \\
\hline
beta & 1.0 \\
\hline
use\_no\_change & True \\
\hline
max\_jsd\_single\_change & 0.01 \\
\hline
max\_frob\_single\_change & 0.0003 \\
\hline
optimize\_already\_fooled & False \\
\hline
use\_disco & False \\
\hline
\end{tabular}
\label{table:HiggsRetrainingTestParameterValues}
\end{table}

\begin{table}[h!]
\caption{Parameter values used for the adversarial attack in the DisCo attack experiment on the Higgs Dataset.}
\begin{tabular}{ |p{3.25cm}|p{3.25cm}| }
\hline
 \multicolumn{2}{|c|}{\textbf{DisCo Attack Hyperparameter Choice}} \\
 \hline
 \textbf{Parameter Name} & \textbf{Value} \\
 \hline \hline
min\_change & 0.015 \\
\hline
step & 0.06 \\
\hline
n\_iterations & 10 \\
\hline
num\_bins & 30 \\
\hline
alpha & 4.0 \\
\hline
beta & 1.0 \\
\hline
use\_no\_change & True \\
\hline
max\_jsd\_single\_change & 0.02 \\
\hline
max\_frob\_single\_change & 0.0003 \\
\hline
optimize\_already\_fooled & False \\
\hline
use\_disco & True \\
\hline
\end{tabular}
\label{table:DiscoAttackParameterValues}
\end{table}

\begin{table}[h!]
\caption{Parameter values used for generating training adversaries in the DisCo detector experiment.}
\begin{tabular}{ |p{3.25cm}|p{3.25cm}| }
\hline
 \multicolumn{2}{|c|}{\textbf{DisCo Detector Attack Hyperparams. (Train)}} \\
 \hline
 \textbf{Parameter Name} & \textbf{Value} \\
 \hline \hline
min\_change & 0.07 \\
\hline
step & 0.08 \\
\hline
n\_iterations & 10 \\
\hline
num\_bins & 30 \\
\hline
alpha & 4.0 \\
\hline
beta & 1.0 \\
\hline
use\_no\_change & True \\
\hline
max\_jsd\_single\_change & 0.07 \\
\hline
max\_frob\_single\_change & 0.0025 \\
\hline
optimize\_already\_fooled & False \\
\hline
use\_disco & True \\
\hline
\end{tabular}
\label{table:DiscoDetectorAttackTrain}
\end{table}

\begin{table}[h!]
\caption{Parameter values used for generating test adversaries in the DisCo detector experiment.}
\begin{tabular}{ |p{3.25cm}|p{3.25cm}| }
\hline
 \multicolumn{2}{|c|}{\textbf{DisCo Detector Attack Hyperparams. (Test)}} \\
 \hline
 \textbf{Parameter Name} & \textbf{Value} \\
 \hline \hline
min\_change & 0.01 \\
\hline
step & 0.05 \\
\hline
n\_iterations & 10 \\
\hline
num\_bins & 30 \\
\hline
alpha & 4.0 \\
\hline
beta & 1.0 \\
\hline
use\_no\_change & True \\
\hline
max\_jsd\_single\_change & 0.01 \\
\hline
max\_frob\_single\_change & 0.00015 \\
\hline
optimize\_already\_fooled & False \\
\hline
use\_disco & True \\
\hline
\end{tabular}
\label{table:DiscoDetectorAttackTest}
\end{table}

\begin{table}[h!]
\caption{Parameter values used for generating training and test adversaries in the for the limited data Pearson correlation experiment.}
\begin{tabular}{ |p{3.25cm}|p{3.25cm}| }
\hline
 \multicolumn{2}{|c|}{\textbf{Limited Data Attack Hyperparameters}} \\
 \hline
 \textbf{Parameter Name} & \textbf{Value} \\
 \hline \hline
min\_change & 0.005 \\
\hline
step & 0.02 \\
\hline
n\_iterations & 10 \\
\hline
num\_bins & 30 \\
\hline
alpha & 6.0 \\
\hline
beta & 1.0 \\
\hline
use\_no\_change & True \\
\hline
max\_jsd\_single\_change & 0.02 \\
\hline
max\_frob\_single\_change & 0.002 \\
\hline
optimize\_already\_fooled & False \\
\hline
use\_disco & False \\
\hline
\end{tabular}
\label{table:LimitedDataDetectorAttackTest}
\end{table}

\subsection{TT vs. WW}

\begin{table}[h!]
\caption{Parameter values used for adversarial attack for the TT vs. WW dataset.}
\begin{tabular}{ |p{3.25cm}|p{3.25cm}| }
\hline
 \multicolumn{2}{|c|}{\textbf{TTvsWW Attack Hyperparameter Choice}} \\
 \hline
 \textbf{Parameter Name} & \textbf{Value} \\
 \hline \hline
min\_change & 0.005 \\
\hline
step & 0.01 \\
\hline
n\_iterations & 10 \\
\hline
num\_bins & 200 \\
\hline
alpha & 6.5 \\
\hline
beta & 1.0 \\
\hline
use\_no\_change & True \\
\hline
max\_jsd\_single\_change & 0.003 \\
\hline
max\_frob\_single\_change & 0.003 \\
\hline
optimize\_already\_fooled & False \\
\hline
use\_disco & False \\
\hline
\end{tabular}
\label{table:TTvsWWAttackParameterValues}
\end{table}

\begin{table}[h!]
\caption{Parameter values used for the augmentation pipeline for the TT vs. WW dataset.}
\begin{tabular}{ |p{3.25cm}|p{3.25cm}| }
\hline
 \multicolumn{2}{|c|}{\textbf{TTvsWW Attack Hyperparameter Choice}} \\
 \hline
 \textbf{Parameter Name} & \textbf{Value} \\
 \hline \hline
min\_change & 0.003 \\
\hline
step & 0.02 \\
\hline
n\_iterations & 10 \\
\hline
num\_bins & 100 \\
\hline
alpha & 6.5 \\
\hline
beta & 1.0 \\
\hline
use\_no\_change & True \\
\hline
max\_jsd\_single\_change & 0.1 \\
\hline
max\_frob\_single\_change & 0.1 \\
\hline
optimize\_already\_fooled & False \\
\hline
use\_disco & False \\
\hline
\end{tabular}
\label{table:TTvsWWAugmentationParameterValues}
\end{table}

\begin{table}[h!]
\caption{Parameter values used for the adversarial detector pipeline for the TT vs. WW dataset for
training and test samples.}
\begin{tabular}{ |p{3.25cm}|p{3.25cm}| }
\hline
 \multicolumn{2}{|c|}{\textbf{TTvsWW Attack Hyperparameter Choice}} \\
 \hline
 \textbf{Parameter Name} & \textbf{Value} \\
 \hline \hline
min\_change & 0.01 \\
\hline
step & 0.01 \\
\hline
n\_iterations & 10 \\
\hline
num\_bins & 100 \\
\hline
alpha & 6.0 \\
\hline
beta & 1.0 \\
\hline
use\_no\_change & False \\
\hline
max\_jsd\_single\_change & 0.01 \\
\hline
max\_frob\_single\_change & 0.03 \\
\hline
optimize\_already\_fooled & False \\
\hline
use\_disco & False \\
\hline
\end{tabular}
\label{table:TTvsWWDetectorParameterValues}
\end{table}

\begin{table}[h!]
\caption{Parameter ranges used for generating train adversaries in the TT vs. WW iterative retraining pipeline (randomized per retraining iteration).}
\begin{tabular}{ |p{3.25cm}|p{3.25cm}| }
\hline
 \multicolumn{2}{|c|}{\textbf{TTvsWW Attack Hyperparam. Ranges (Train)}} \\
 \hline
 \textbf{Parameter Name} & \textbf{Value / Range} \\
 \hline \hline
min\_change & [0.001, 0.005] (uniform) \\
\hline
step & [0.005, 0.03] (uniform) \\
\hline
n\_iterations & 10 \\
\hline
num\_bins & 100 \\
\hline
alpha & [3.0, 10.0] (uniform) \\
\hline
beta & [0.5, 2.5] (uniform) \\
\hline
use\_no\_change & True \\
\hline
max\_jsd\_single\_change & [0.005, 0.03] (uniform) \\
\hline
max\_frob\_single\_change & [0.0001, 0.001] (uniform) \\
\hline
optimize\_already\_fooled & False \\
\hline
use\_disco & False \\
\hline
\end{tabular}
\label{table:TTvsWWRetrainParameterRangesTrain}
\end{table}

\begin{table}[h!]
\caption{Parameter values used for generating test adversaries in the TT vs. WW iterative retraining pipeline.}
\begin{tabular}{ |p{3.25cm}|p{3.25cm}| }
\hline
 \multicolumn{2}{|c|}{\textbf{TTvsWW Attack Hyperparameters (Test)}} \\
 \hline
 \textbf{Parameter Name} & \textbf{Value} \\
 \hline \hline
min\_change & 0.002 \\
\hline
step & 0.005 \\
\hline
n\_iterations & 10 \\
\hline
num\_bins & 100 \\
\hline
alpha & 8.0 \\
\hline
beta & 1.0 \\
\hline
use\_no\_change & True \\
\hline
max\_jsd\_single\_change & 0.01 \\
\hline
max\_frob\_single\_change & 0.0001 \\
\hline
optimize\_already\_fooled & False \\
\hline
use\_disco & False \\
\hline
\end{tabular}
\label{table:TTvsWWRetrainParameterValuesTest}
\end{table}

\subsection{Donut}

\begin{table}[h!]
\caption{Parameter values used for generating test adversaries in the DonutDummy experiment.}
\begin{tabular}{ |p{3.25cm}|p{3.25cm}| }
\hline
 \multicolumn{2}{|c|}{\textbf{DonutDummy Attack Hyperparameters (Test)}} \\
 \hline
 \textbf{Parameter Name} & \textbf{Value} \\
 \hline \hline
min\_change & 0.001 \\
\hline
num\_candidates & 150 \\
\hline
n\_iterations & 10 \\
\hline
num\_bins & 70 \\
\hline
alpha & 6.0 \\
\hline
beta & 1.0 \\
\hline
use\_no\_change & True \\
\hline
max\_jsd\_single\_change & 0.005 \\
\hline
max\_frob\_single\_change & 0.05 \\
\hline
optimize\_already\_fooled & False \\
\hline
use\_disco & False \\
\hline
\end{tabular}
\label{table:DonutDummyAttackParameterValues}
\end{table}

\begin{table}[h!]
\caption{Parameter values used for generating training and test adversaries in the DonutDummy adversarial detector pipeline experiment.}
\begin{tabular}{ |p{3.25cm}|p{3.25cm}| }
\hline
 \multicolumn{2}{|c|}{\textbf{DonutDummy Detector Attack Hyperparams.}} \\
 \hline
 \textbf{Parameter Name} & \textbf{Value} \\
 \hline \hline
min\_change & 0.001 \\
\hline
num\_candidates & 150 \\
\hline
n\_iterations & 10 \\
\hline
num\_bins & 60 \\
\hline
alpha & 6.0 \\
\hline
beta & 1.0 \\
\hline
use\_no\_change & True \\
\hline
max\_jsd\_single\_change & 0.005 \\
\hline
max\_frob\_single\_change & 0.05 \\
\hline
optimize\_already\_fooled & False \\
\hline
use\_disco & False \\
\hline
\end{tabular}
\label{table:DonutDummyDetectorPipelineAttackParams}
\end{table}

\end{appendices}

\end{document}